\documentclass[final,5p,times,twocolumn,authoryear]{elsarticle}

\usepackage{amssymb}
\usepackage{amsmath}
\usepackage{tabularx}
\usepackage{array}
\usepackage{booktabs}
\usepackage{booktabs}
\usepackage{multirow}
\usepackage{tabularx}
\usepackage{array}
\usepackage{graphicx}
\usepackage{dblfloatfix}

\usepackage{hyperref}

\hypersetup{
	colorlinks=true,
	linkcolor=blue,
	citecolor=blue,
	urlcolor=blue
}

\newcolumntype{Y}{>{\raggedright\arraybackslash}X}

\journal{Expert Systems With Applications}

\begin{document}

\begin{frontmatter}



\title{EEGProbe: Do Vision-Language Models Really Understand EEG? A Benchmark for Signal-Grounded EEG Signals Evaluation}

\author[label1]{Haozhe Zhu\fnref{equal}} 
\ead{zhu.haozhe.305@s.kyushu-u.ac.jp}
\author[label1]{Dan Yuan\fnref{equal}} 
\ead{yuan.dan.718@s.kyushu-u.ac.jp}
\author[label1]{Chenghao Gu} 
\ead{gu.chenghao.564@s.kyushu-u.ac.jp}
\author[label1]{Feike Xu} 
\ead{xu.feike.383@s.kyushu-u.ac.jp}
\author[label2]{Yue Wen\corref{cor2}} 
\ead{wenyue@suse.edu.cn}
\author[label3]{Shin'ichi Konomi\corref{cor1}}
\ead{konomi.shinichi.848@m.kyushu-u.ac.jp}

\fntext[equal]{These authors contributed equally to this work.}

\cortext[cor1]{Corresponding author.}
\cortext[cor2]{Corresponding author.}

\affiliation[label1]{organization={Graduate School of Information Science and Electrical Engineering, Kyushu University},
            city={Fukuoka},
            postcode={819-0395}, 
            country={Japan}}

\affiliation[label2]{organization={Department of Civil Engineering, Sichuan University of Science and Engineering},
	city={Zigong},
	postcode={643002}, 
	country={China}}

\affiliation[label3]{organization={Faculty of Arts and Science, Kyushu University},
	city={Fukuoka},
	postcode={819-0395}, 
	country={Japan}}

\begin{abstract}
Vision-language models (VLMs) have shown strong multimodal capabilities across scientific and biomedical domains, yet their ability to understand electroencephalography (EEG) signals remains unclear. We introduce EEGProbe, a signal-grounded benchmark for evaluating whether general-purpose VLMs can extract and reason from minimally processed, channel-wise EEG without task-specific feature engineering or EEG-specific adaptation. EEGProbe contains 620 EEG-centered questions across 13 subtasks and three capability categories: EEG Signal Classification, EEG Cross-modal Alignment, and EEG Neurophysiological Reasoning. To reduce shortcut-based answering, EEGProbe preserves the temporal structure of the recordings where possible, minimizes benchmark-induced processing, and applies signal-necessity testing followed by expert verification.
We evaluate 11 representative open- and closed-source VLMs under a unified zero-shot protocol. The strongest model, GPT-5.6 Sol, achieves an overall score of only 41.63\%, with cross-modal EEG alignment emerging as a major bottleneck. We further examine model behavior from a neurophysiological perspective and analyze how temporal, spectral, spatial, and physiological EEG structure relates to model successes and failures. Inference-time neurophysiological guidance can reshape model predictions by helping models recognize the neurophysiological patterns present in the input and connect them with relevant neurological knowledge, narrowing the gap between general multimodal competence and EEG understanding.
EEGProbe provides a testbed for current VLMs and a basis for developing future multimodal models that can interpret EEG evidence more reliably.
The EEGProbe benchmark will be publicly available at \url{https://huggingface.co/datasets/2122ZHU/EEGProbe}.
	
\end{abstract}



\begin{keyword}
	Electroencephalography (EEG) \sep
	Vision-language models(VLMs) \sep
	Benchmark \sep
	Multimodal reasoning \sep
	Neural-signal understanding \sep
	Brain--computer interfaces
\end{keyword}

\end{frontmatter}



\section{Introduction}

\begin{figure*}[t]
	\centering
	\includegraphics[width=\textwidth]{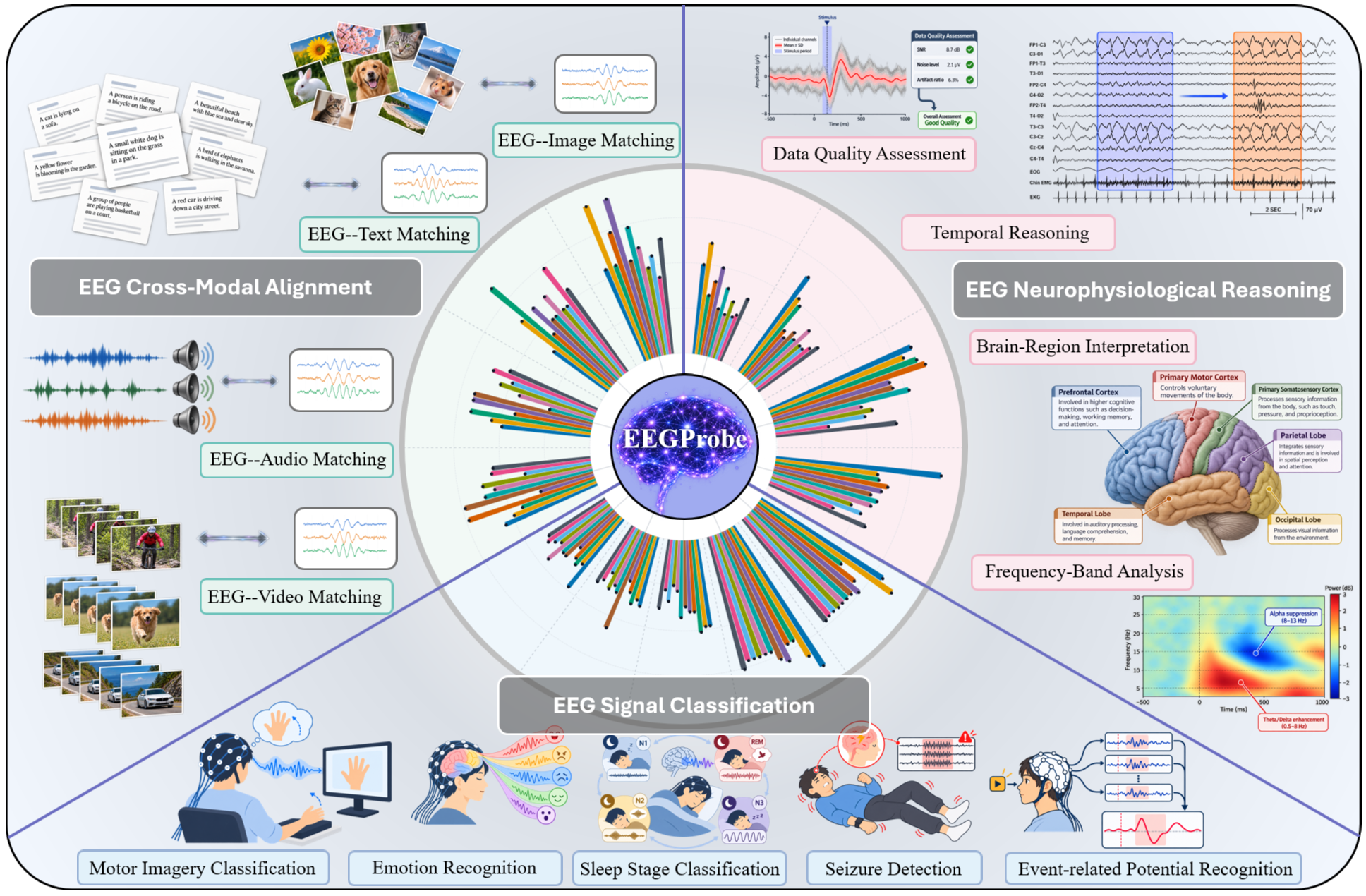}
	\caption{
		Overview of EEGProbe. The benchmark evaluates general-purpose vision-language models across three categories of EEG understanding: EEG Signal Classification, EEG Cross-modal Alignment, and EEG Neurophysiological Reasoning. EEGProbe contains 13 subtasks covering motor imagery, emotion, sleep, seizure, ERP, cross-modal matching with image, text, audio, and video stimuli, and reasoning about frequency, brain regions, temporal structure, and signal quality.
	}
	\label{fig:eegprobe_overview}
\end{figure*}

Vision-language models (VLMs)~\citep{radford2021learning,zhang2024vision} have developed strong capabilities for interpreting visual and semantic information across a wide range of domains. Whether these capabilities extend to electrophysiological signals such as electroencephalography (EEG), however, remains much less understood. Unlike natural images or text, EEG encodes information through multichannel electrical activity whose interpretation depends jointly on temporal, spatial, and spectral structure. This raises an important question for neural engineering: \textit{can a general-purpose VLM extract meaningful information from EEG signals without being explicitly trained for EEG?}
EEG is a non-invasive measure of brain electrical activity and supports applications including brain--computer interfaces (BCIs), cognitive and affective monitoring, sleep analysis, and neurological assessment~\citep{liu2025recent,chen2025eeg}. Most existing EEG systems are designed for a specific task. Motor imagery decoding~\citep{al2021deep,padfield2019eeg,zhao2025tcanet}, emotion recognition~\citep{jenke2014feature,jafari2023emotion}, sleep staging~\citep{aboalayon2016sleep,phan2022automatic}, seizure detection~\citep{boonyakitanont2020review,ein2023eeg}, and event-related potential (ERP) analysis~\citep{light2010electroencephalography,fietz2025empathy} typically rely on dedicated preprocessing, representations, and task-specific models. These methods are effective for their intended applications, but they address a different question from the one considered here: rather than asking how well a specialized model can solve a particular EEG task, we ask what a broadly trained multimodal model can infer from EEG signals without task-specific optimization.
This question is difficult because EEG signals are noisy, non-stationary, and highly heterogeneous across subjects, recording systems, and experimental paradigms. Its informative structure may appear as transient event-related responses, sustained oscillations, distributed spatial patterns, or longer-term physiological dynamics~\citep{yao2025foundation}. Conventional EEG decoding therefore relies heavily on signal processing and representation learning to expose task-relevant structure before prediction~\citep{schirrmeister2017deep}. Recent EEG foundation models further improve transferability through dedicated neural encoders, large-scale pretraining, data standardization, and downstream adaptation~\citep{liu2026eeg}. Although these approaches improve transferability and downstream EEG performance, they make it difficult to determine how much electrophysiological evidence a general-purpose model can extract without EEG-specific adaptation.

This distinction is particularly important when evaluating VLMs. \textit{A correct answer to an EEG-related question does not necessarily mean that the model has used the EEG signal.} A model may instead rely on textual cues, candidate answers, prior knowledge, or representations in which task-relevant information has already been extracted. Evaluating EEG understanding therefore requires not only measuring whether an answer is correct, but also establishing whether the answer depends on the EEG evidence itself. It is equally important to limit task-specific signal processing that could solve part of the problem before the input reaches the model.

The amount of EEG-specific guidance provided at inference time can also affect
the apparent capability of a model. Instructions about relevant frequency
bands, brain regions, or expected physiological patterns may improve
predictions, but they also make it harder to distinguish the model's own
signal-understanding ability from knowledge supplied by the user. EEGProbe
therefore minimizes task-specific guidance and evaluates how much
task-relevant information general-purpose VLMs can recover from minimally
processed EEG under a unified zero-shot setting.
To study this problem, we introduce EEGProbe, a signal-grounded benchmark for evaluating general-purpose VLMs on minimally processed, channel-wise EEG (Figure~\ref{fig:eegprobe_overview}). EEGProbe organizes EEG understanding into three complementary categories. EEG Signal Classification evaluates recognition of cognitive, affective, physiological, and neurological states, including motor imagery, emotion, sleep stages, seizures, and ERPs. EEG Cross-modal Alignment evaluates whether neural responses can be associated with visual, textual, auditory, and video stimuli. EEG Neurophysiological Reasoning examines whether models can interpret spectral, spatial, temporal, and signal-quality evidence. Together, the 13 subtasks assess whether VLMs can recognize information encoded in EEG, relate neural responses to external stimuli, and reason from the neurophysiological structure of the signal.
A central design principle of EEGProbe is to preserve the electrophysiological evidence while minimizing benchmark-induced processing. Channel-wise temporal structure is retained whenever possible, and no handcrafted answer-related features or predictions from dedicated EEG decoders are provided. We further introduce a signal-necessity test: candidate questions are presented without their EEG input, and any question that can still be answered correctly is removed. The remaining instances are then verified by domain experts for answer correctness and EEG grounding. This procedure is designed to separate reasoning \textit{about EEG} from reasoning \textit{from EEG}.

The resulting benchmark contains 620 EEG-centered question--answer pairs spanning 13 subtasks. We evaluate 11 representative open- and closed-source VLMs under a unified zero-shot protocol. The strongest evaluated model, GPT-5.6 Sol, achieves an overall score of only 41.63\%, and cross-modal EEG alignment remains close to chance level across all four modalities. Further analysis reveals substantial differences in task-level capability profiles and shows that neurophysiological guidance can reshape model predictions and improve performance on selected tasks. These findings suggest that general multimodal competence does not yet translate into reliable use of electrophysiological evidence.
Beyond comparing existing models, EEGProbe provides a framework for studying how future multimodal models might better interact with neural signals. By analyzing performance across temporal, spectral, spatial, and physiological EEG structures, the benchmark helps identify where current models fail and which forms of neurophysiological guidance can improve their use of EEG evidence. This provides a basis for developing models and inference strategies that depend less on task-specific feature engineering and more directly exploit the structure present in electrophysiological recordings.

Our main contributions are summarized as follows:

\begin{itemize}
	
	\item We introduce \textbf{EEGProbe}, a signal-grounded benchmark comprising 620 questions and 13 subtasks across three dimensions of EEG understanding: \textbf{EEG Signal Classification}, \textbf{EEG Cross-modal Alignment}, and \textbf{EEG Neurophysiological reasoning}.
	
	\item We develop an evaluation protocol that minimizes benchmark-induced signal processing and combines \textbf{signal-necessity testing} with expert verification, reducing the possibility that models succeed without using the EEG evidence.
	
	\item We evaluate \textbf{11 representative open- and closed-source VLMs} under a unified zero-shot setting and reveal substantial limitations in current EEG understanding. Further neurophysiological analysis shows that task-relevant guidance can reshape model predictions and improve performance on selected tasks, providing directions for future EEG-capable multimodal models.
	
\end{itemize}

\section{Related Work}

\subsection{EEG Decoding and Representation Learning}

EEG decoding has long depended on representations that make task-relevant structure easier to identify from noisy recordings. Deep learning has reduced the need for manually designed features by learning these representations directly from data. Convolutional neural networks (CNNs)~\citep{he2016deep} have played an important role in this transition. EEGNet, for example, uses temporal and spatial convolutions to capture frequency-related patterns and inter-channel dependencies in a compact architecture that can be applied across several BCI paradigms~\citep{lawhern2018eegnet}. Later CNN-based methods introduced multi-scale temporal filters, spatial attention, and feature fusion to improve EEG feature extraction, although their receptive fields still tend to favor relatively local temporal and spatial structure.

Transformer-based models address part of this limitation by using self-attention to model longer-range dependencies. EEG Conformer~\citep{song2022eeg} combines convolutional feature extraction with Transformer layers, allowing local temporal--spatial patterns and more global dependencies to be modeled within the same architecture. Subsequent work has explored multi-scale, hierarchical, and spatial--temporal attention to capture interactions across channels and over longer temporal ranges.

Riemannian approaches provide another way to represent multichannel EEG by modeling covariance structure directly. In these methods, each trial is commonly represented as a symmetric positive-definite (SPD) matrix, so that second-order relationships among channels are preserved on a non-Euclidean manifold. Recent work has combined this geometric representation with deep neural networks to learn task-relevant transformations of SPD features~\citep{wilson2025deep}. Other methods further introduce manifold-aware spatial transformations and graph structures to model organized relationships among EEG channels~\citep{tibermacine2026stiefel}.

Although these approaches differ in architecture and representation, they share the goal of learning EEG features that are useful for a particular decoding task. EEGProbe considers a different setting. Instead of introducing another EEG-specific representation or decoder, we evaluate whether general-purpose VLMs can extract useful evidence from minimally processed EEG without task-specific feature engineering. EEGProbe instead examines what general-purpose multimodal models can infer directly from minimally processed EEG.

\subsection{Foundation and Multimodal Models for EEG Understanding}

Foundation models have recently been introduced into EEG analysis to learn representations that can transfer across downstream tasks. Neuro-GPT combines an EEG encoder with a GPT-based architecture and uses self-supervised reconstruction of masked EEG segments before fine-tuning for motor imagery classification~\citep{cui2024neuro}. Other EEG foundation models similarly use self-supervised or generative pretraining to improve transfer across subjects, datasets, and paradigms. These methods aim to learn reusable EEG representations, but the models are still explicitly trained to capture EEG-specific structure.

Another line of work connects EEG with semantic modalities such as language and vision. Thought2Text maps visually evoked EEG to semantic representations and adapts instruction-following language models to generate textual descriptions from EEG-derived embeddings~\citep{mishra2025thought2text}. MindAlign further aligns EEG, visual, and linguistic representations for zero-shot visual decoding~\citep{chen2026mindalign}, while EEG-VLM applies a vision-language architecture to EEG-based sleep-stage prediction~\citep{qiu2026eeg}. These studies show that foundation models can be adapted to link electrophysiological activity with semantic information.

Most existing approaches rely on EEG-specific encoders, pretraining, cross-modal alignment, or downstream adaptation. Their performance therefore reflects both the underlying model and the EEG-specific design. This makes it difficult to determine what a general-purpose multimodal model can extract from EEG without such support.
EEGProbe addresses this by evaluating existing general-purpose VLMs on minimally processed EEG, without EEG-specific training, task-specific feature engineering, or a dedicated EEG encoder. This separation is necessary to test how far general multimodal capabilities transfer to electrophysiological signals and where EEG-specific modeling is still needed.

\subsection{Benchmarks for Scientific and Biomedical Foundation Models}

Scientific benchmark design has increasingly shifted from general knowledge recall toward reasoning over domain-specific evidence. GPQA~\citep{rein2023gpqa} evaluates graduate-level reasoning in biology, physics, and chemistry, while MMMU~\citep{yue2024mmmu} extends multimodal evaluation across a broad range of academic disciplines and visual inputs. LAB-Bench~\citep{laurent2024lab} further targets practical biology research tasks involving scientific figures, literature, databases, and biological sequences.
A similar trend has emerged in biomedical evaluation. MultiMedQA~\citep{singhal2023large} focuses primarily on medical knowledge and question answering, whereas OmniMedVQA~\citep{hu2024omnimedvqa} evaluates models across diverse medical imaging modalities and anatomical regions. MedXpertQA~\citep{zuo2025medxpertqa} places greater emphasis on expert-level reasoning over clinically grounded textual and multimodal evidence. More recent benchmarks have begun to include physiological time series. HEARTS~\citep{li2026hearts} evaluates reasoning across multiple modalities, including EEG, ECG, PPG, EMG, and EOG, while EEG-specific benchmarks compare pretrained EEG foundation models across datasets and downstream paradigms~\citep{xiong2025eeg}. These benchmarks extend foundation-model evaluation beyond language and images, but they mainly assess either general time-series reasoning or models that have been explicitly trained to learn signal-specific representations.
BrainBench~\citep{zhou2026brainbench} moves closer to the setting considered here by evaluating how language-model agents solve EEG problems through executable signal-processing workflows. EEGProbe focuses on a different question: whether a general-purpose VLM can extract useful evidence from minimally processed EEG without an EEG-specific encoder, task-dependent feature engineering, or an external signal-processing pipeline. We therefore evaluate models directly on the EEG input and explicitly test whether the signal is necessary for answering each question, separating reasoning \textit{from EEG signals} from reasoning merely \textit{about EEG signals}.

\section{Methods}
\label{sec:methods}

\begin{figure*}[!t]
	\centering
	\includegraphics[width=\textwidth]{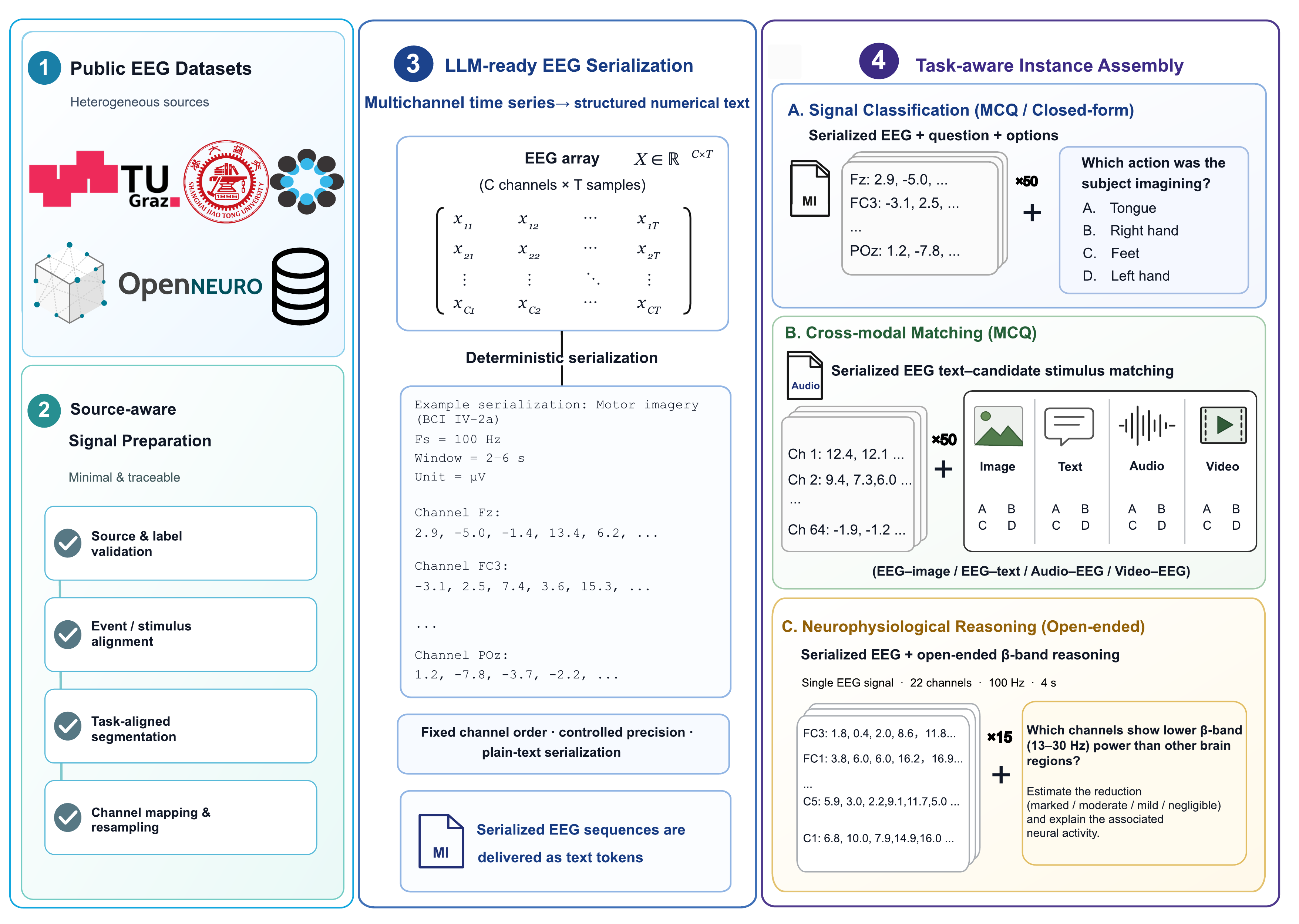}
	\caption{
		EEG data preparation and serialization pipeline in EEGProbe.
		EEG recordings undergo minimal source-aware preparation, followed by deterministic serialization of multichannel time series into structured numerical text.
		The serialized signals are then assembled into task-specific instances for the three benchmark dimensions.
	}
	\label{fig:eeg_serialization_pipeline}
\end{figure*}

EEGProbe is designed to evaluate general-purpose VLMs across heterogeneous EEG paradigms while limiting task-specific processing of the input signal. Figure~\ref{fig:eeg_serialization_pipeline} summarizes the preparation and serialization pipeline used to construct model-ready EEG instances.

\subsection{General Criteria}

EEGProbe is guided by the following four criteria:

\noindent\textbf{EEG necessity.}
The correct answer should depend on the EEG input rather than on textual cues, candidate options, or prior knowledge alone. We therefore retain only questions that pass the signal-necessity test described in Section~\ref{Dataset Construction}.

\noindent\textbf{Signal fidelity.}
EEG inputs should preserve the channel-wise temporal structure of the source recordings as much as possible. Benchmark-side processing is therefore limited to task-required operations such as segmentation, necessary resampling or alignment, predefined channel handling, numerical serialization, and rounding. We avoid handcrafted features, spectral summaries, classifier outputs, or other transformations that explicitly expose answer-related information.

\noindent\textbf{Objectivity.}
Scoring procedures are fixed before evaluation and applied consistently across models. Multiple-choice and closed-form responses are evaluated using predefined answer rules, while open-ended responses are assessed against expert-written reference answers and fixed rubrics.

\noindent\textbf{Stability.}
All models are evaluated using the same frozen benchmark instances, prompts, answer rules, and scoring procedures. For open-ended responses, standardized judging and expert verification are used to reduce variability in semantic evaluation.

These criteria guide the dataset construction and evaluation design described in the following sections.

\subsection{Dataset Construction}
\label{Dataset Construction}

\begin{figure*}[t]
	\centering
	\includegraphics[width=\textwidth]{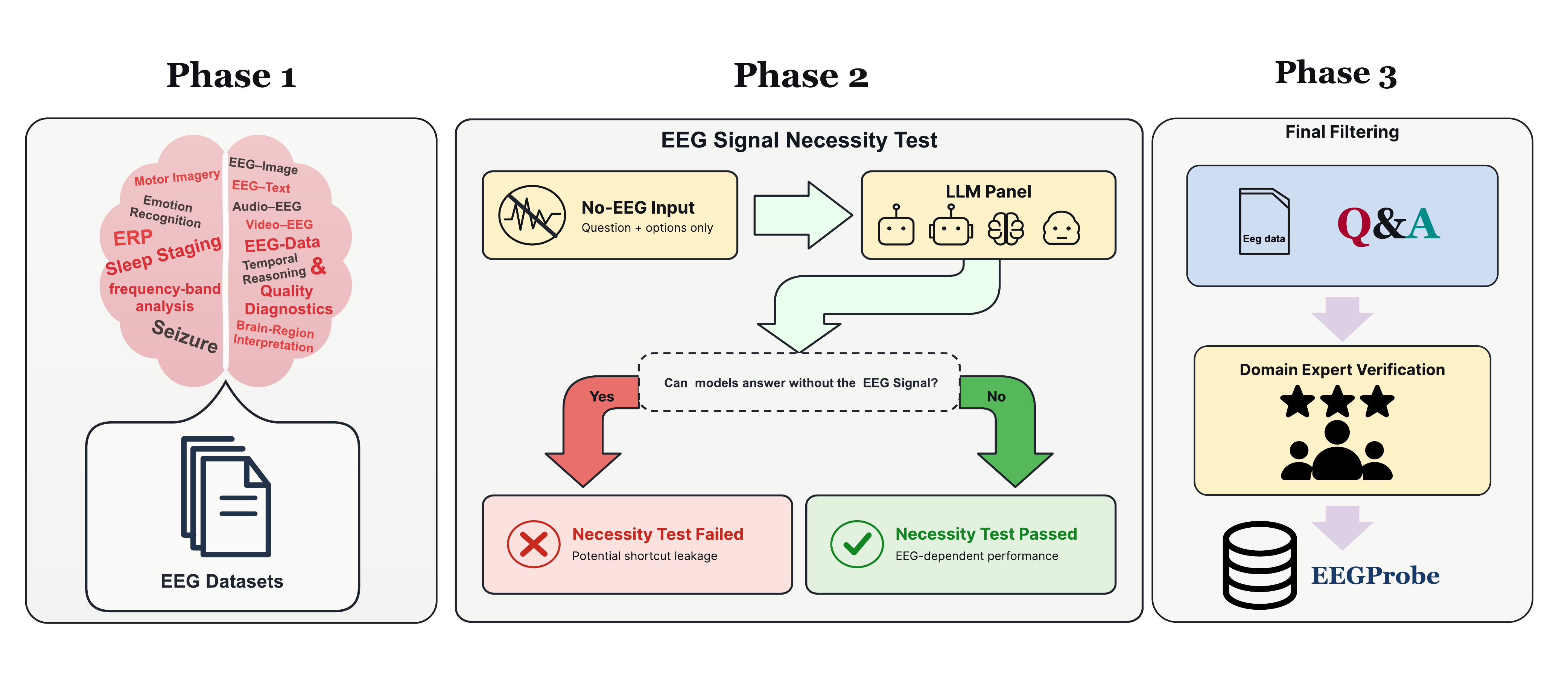}
	\caption{
		Construction pipeline of EEGProbe.
		Candidate instances are collected from multiple EEG datasets and tasks
		(Phase 1), filtered through an EEG signal-necessity test
		(Phase 2), and then verified by domain experts for answer correctness
		and EEG grounding (Phase 3).
	}
	\label{fig:data_construction}
\end{figure*}

As shown in Figure~\ref{fig:data_construction}, EEGProbe is constructed in three stages: data collection, signal-necessity testing, and expert verification.

\noindent\textbf{Data collection.}
We curate EEG recordings from 15 publicly released or research-accessible datasets covering a range of experimental and clinical paradigms.
For signal classification, we use BCI Competition IV 2a for motor imagery
\citep{tangermann2012bci}; ERP CORE P3 and the Wadsworth P300 dataset
(BCI Competition III Dataset II) for ERP recognition
\citep{kappenman2021erpcore,blankertz2006bci}; SEED and SEED-IV for
emotion recognition \citep{zheng2015seed,zheng2019emotionmeter};
DOD-H and Sleep-EDFx for sleep staging
\citep{guillot2020dreem,kemp2000sleep}; and CHB-MIT for seizure detection
\citep{guttag2010chbmit}.
For EEG Cross-modal Alignment, we use SparrKULee for auditory stimuli
\citep{accou2024sparrkulee}, THINGS-EEG2 for images
\citep{gifford2022things}, ZuCo 1.0 for natural-language stimuli
\citep{hollenstein2018zuco}, and SEED-DV for video stimuli
\citep{liu2024eeg2video}.
Neurophysiological Reasoning tasks additionally use ERP CORE, EEGMMIDB,
TUAR v3.0.1, the CAP Sleep Database, BCI Competition IV 2a, and Sleep-EDFx
\citep{kappenman2021erpcore,schalk2004bci2000,obeid2016tuh,
	terzano2001cap,tangermann2012bci,kemp2000sleep}.
A source dataset may contribute to more than one subtask when its annotations
and signal characteristics support different forms of evaluation.

For each dataset, we retain the signal recordings, event annotations, channel
metadata, labels, and paired stimuli required by the corresponding task.
Benchmark-side processing is limited to task-aligned segmentation, necessary
resampling or temporal alignment, predefined channel handling, and numerical
serialization. Channel-wise time-domain structure is preserved whenever
possible, and samples are reconstructed from the original recordings when the
available processed version does not meet this minimal-processing policy.

\noindent\textbf{Signal-necessity testing.}
To remove questions that can be answered without EEG evidence, we conduct a
signal-necessity test. For each candidate question--answer pair, the EEG input
is removed and the remaining question is presented to four language models:
ChatGPT~\citep{openai2022chatgpt}, Qwen~\citep{bai2023qwen},
Claude~\citep{anthropic2024claude3}, and Grok~\citep{xai2023grok}.
If any model can correctly answer the question without access to the EEG,
the instance is excluded. This step reduces the influence of textual cues,
candidate-option biases, and prior knowledge unrelated to the signal.

\noindent\textbf{Expert verification.}
The remaining instances are manually reviewed by two authors with research
experience in EEG signal analysis and neurophysiological interpretation.
They verify both the ground-truth answer and whether the EEG provides sufficient
evidence to support it. Only instances that satisfy both criteria are retained
in the final benchmark.

\subsection{Evaluation Dimensions}
\label{Evaluation Dimensions}

\begin{figure*}[!t]
	\centering
	\includegraphics[width=0.82\textwidth]{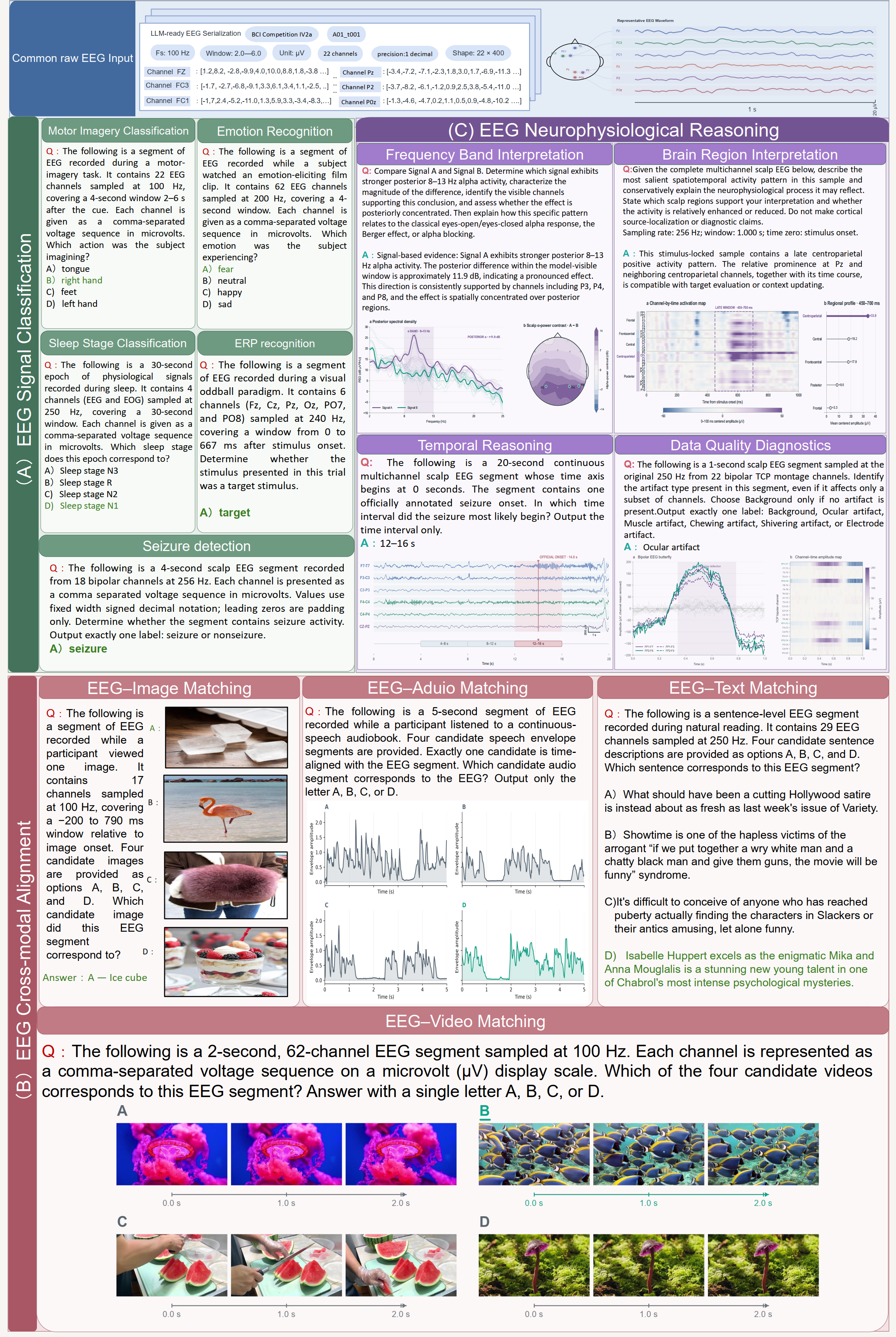}
	\caption{
		Representative task examples in EEGProbe.
		(a) EEG Signal Classification evaluates state and response recognition from EEG.
		(b) EEG Cross-modal Alignment assesses correspondence between EEG responses and external visual, textual, auditory, and video stimuli.
		(c) EEG Neurophysiological Reasoning evaluates spectral, spatial, temporal, and signal-quality interpretation.
	}
	\label{fig:task_examples}
\end{figure*}

EEGProbe organizes 13 subtasks into three evaluation categories, as illustrated
in Figure~\ref{fig:task_examples}.

\begin{enumerate}
	
	\item \textbf{EEG Signal Classification} (Figure~\ref{fig:task_examples}(A)) evaluates whether models can infer
	cognitive, affective, physiological, and neurological states or responses
	from EEG. It comprises motor imagery classification, emotion recognition,
	sleep stage classification, seizure detection, and ERP recognition.
	
	\item \textbf{EEG Cross-modal Alignment} (Figure~\ref{fig:task_examples}(B)) evaluates whether models can match
	EEG responses with the external stimuli that elicited them. It comprises
	EEG--image matching, EEG--text matching, EEG--audio matching, and
	EEG--video matching.
	
	\item \textbf{EEG Neurophysiological Reasoning} (Figure~\ref{fig:task_examples}(C)) evaluates whether models can
	interpret signal properties beyond task-label prediction. It comprises
	frequency-band analysis, brain-region interpretation, temporal reasoning,
	and data quality assessment.
	
\end{enumerate}

The tasks are presented using three response formats according to their
evaluation requirements: multiple-choice, closed-form, and open-ended
responses. Question formats are selected according to each subtask's response requirements
and may differ within the same evaluation category (Table~\ref{tab:question_formats}).

\begin{table*}[!t]
	\centering
	\caption{Question formats used in EEGProbe.}
	\label{tab:question_formats}
	
	\footnotesize
	\renewcommand{\arraystretch}{1.25}
	\setlength{\tabcolsep}{6pt}
	
	\begin{tabularx}{\textwidth}{
			@{}
			>{\raggedright\arraybackslash}p{0.18\textwidth}
			Y
			Y
			@{}
		}
		\toprule
		\textbf{Question Format}
		&
		\textbf{Response Requirement}
		&
		\textbf{Representative EEG Use}
		\\
		\midrule
		
		\textbf{Multiple-choice}
		&
		Models select the correct answer from four candidate options.
		&
		EEG Signal Classification and Cross-modal Alignment.
		\\
		
		\addlinespace[3pt]
		
		\textbf{Closed-form}
		&
		Models provide a constrained response, such as a category,
		numerical value, or short phrase.
		&
		Seizure detection, ERP recognition, temporal reasoning,
		and data quality assessment.
		\\
		
		\addlinespace[3pt]
		
		\textbf{Open-ended response}
		&
		Models provide a free-form interpretation grounded in the given EEG signal.
		&
		Frequency-band analysis and brain-region interpretation.
		\\
		
		\bottomrule
	\end{tabularx}
\end{table*}

\subsection{Question Formats and Scoring}

EEGProbe contains 620 instances across 13 subtasks and three evaluation categories. The subtasks use three question formats: multiple-choice, closed-form, and open-ended. Their composition is summarized in Table~\ref{tab:task_dimensions}.

\begin{table*}[t]
	\centering
	\caption{Task composition and question formats in EEGProbe.}
	\label{tab:task_dimensions}
	\small
	\renewcommand{\arraystretch}{1.15}
	\begin{tabularx}{\textwidth}{
			@{}
			>{\raggedright\arraybackslash}p{3.8cm}
			c
			>{\raggedright\arraybackslash}X
			c
			>{\centering\arraybackslash}p{2.5cm}
			@{}
		}
		\toprule
		\textbf{Category} &
		\textbf{Abbr.} &
		\textbf{Subtask} &
		\textbf{\# Instances} &
		\textbf{Question Format} \\
		\midrule
		
		\multirow{5}{3.8cm}{EEG Signal Classification}
		& MI  & Motor Imagery Classification        & 50 & Multiple-choice \\
		& ER  & Emotion Recognition                 & 50 & Multiple-choice \\
		& SSC & Sleep Stage Classification          & 50 & Multiple-choice \\
		& SD  & Seizure Detection                   & 50 & Closed-form \\
		& ERP & Event-Related Potential Recognition & 50 & Closed-form \\
		
		\midrule
		
		\multirow{4}{3.8cm}{EEG Cross-Modal Alignment}
		& EIM & EEG--Image Matching & 50 & Multiple-choice \\
		& ETM & EEG--Text Matching  & 50 & Multiple-choice \\
		& EAM & EEG--Audio Matching & 50 & Multiple-choice \\
		& EVM & EEG--Video Matching & 50 & Multiple-choice \\
		
		\midrule
		
		\multirow{4}{3.8cm}{EEG Neurophysiological Reasoning}
		& FBA & Frequency-Band Analysis     & 50 & Open-ended \\
		& BRI & Brain-Region Interpretation & 37 & Open-ended \\
		& TR  & Temporal Reasoning           & 33 & Closed-form \\
		& DQA & Data Quality Assessment      & 50 & Closed-form \\
		
		\midrule
		\multicolumn{3}{l}{\textbf{Total}} &
		\textbf{620} &
		-- \\
		\bottomrule
	\end{tabularx}
\end{table*}

Multiple-choice and closed-form responses are scored against predefined gold answers. For multiple-choice questions, we extract the final valid option.
For closed-form questions, the final explicit prediction is matched to the corresponding answer. Ambiguous responses are counted as incorrect.
For open-ended questions, reference answers are prepared by authors with research experience in EEG analysis. Each reference answer is based on the model-visible EEG and the corresponding task annotation, and specifies the expected finding, supporting EEG evidence, neurophysiological interpretation, and relevant limitations when needed. These elements are converted into item-specific weighted rubrics, with higher weights assigned to criteria that are more important for a correct and complete answer.
The rubric weights are initialized uniformly across scoring criteria and then adjusted by the expert reviewers based on the relative importance of each criterion to the neurophysiological interpretation required by the question.
Representative complete examples for FBA and BRI are provided in~\ref{app:rubric_examples}.

\subsection{Benchmark Characteristics}

EEGProbe contains 620 EEG-centered question--answer pairs across three evaluation categories and 13 subtasks. Beyond task diversity, the benchmark has three characteristics that are important for evaluating general-purpose models on EEG.

\begin{enumerate}
	
	\item \textbf{EEG-specific signal interpretation.}
	Many tasks require models to combine the observed EEG evidence with domain knowledge about sensorimotor rhythms, sleep-related activity, epileptic patterns, event-related potentials, regional channel organization, and physiological artifacts. The benchmark therefore evaluates not only whether models know relevant EEG concepts, but also whether they can apply that knowledge to the signal presented in each instance.
	
	\item \textbf{Substantial signal-domain shift.}
	EEG differs markedly from the language and visual data on which general-purpose models are commonly trained. It consists of multichannel, low-signal-to-noise time series that vary across subjects, recording systems, channel configurations, and experimental paradigms. In EEGProbe, the signal is primarily presented as channel-wise temporal numerical sequences without handcrafted answer-related features, creating a challenging zero-shot setting for general-purpose models.
	
	\item \textbf{Broad coverage of EEG capabilities and paradigms.}
	EEGProbe combines EEG Signal Classification, EEG Cross-modal Alignment, and EEG Neurophysiological Reasoning across 13 subtasks drawn from diverse experimental and clinical settings. The benchmark includes multiple-choice, closed-form, and open-ended responses, covering state recognition, stimulus matching, and signal interpretation within a single evaluation framework. This design allows model behavior to be compared across heterogeneous EEG tasks rather than within a single decoding problem.
	
\end{enumerate}

\section{Experiments}

\subsection{Experimental Setup}
\label{ES}

We evaluate 11 representative VLMs on the frozen EEGProbe benchmark under a unified zero-shot setting, without EEG-specific training, fine-tuning, or task adaptation. All models are evaluated on the same instances using identical task inputs, prompts, answer options, reference answers, and scoring criteria.

\paragraph{Multiple-Choice and Closed-Form Questions}

Each multiple-choice question contains four candidate options with one correct
answer. Closed-form questions require a constrained response, such as a
predefined category, numerical value, or short phrase.
Because some models produce explanations or return answers in formats that
differ from the requested output, we use a deterministic parsing procedure
rather than exact matching over the full response. For multiple-choice
questions, the parser extracts the model's final valid option. For closed-form
questions, the final explicit prediction is mapped to the corresponding
predefined answer when possible. Explanatory text is ignored as long as the
response contains a unique final answer. Responses that cannot be mapped
unambiguously, or that contain contradictory final predictions, are marked as
incorrect.

\paragraph{Open-Ended Questions}

\begin{figure*}[!t]
	\centering
	\includegraphics[
	width=\textwidth
	]{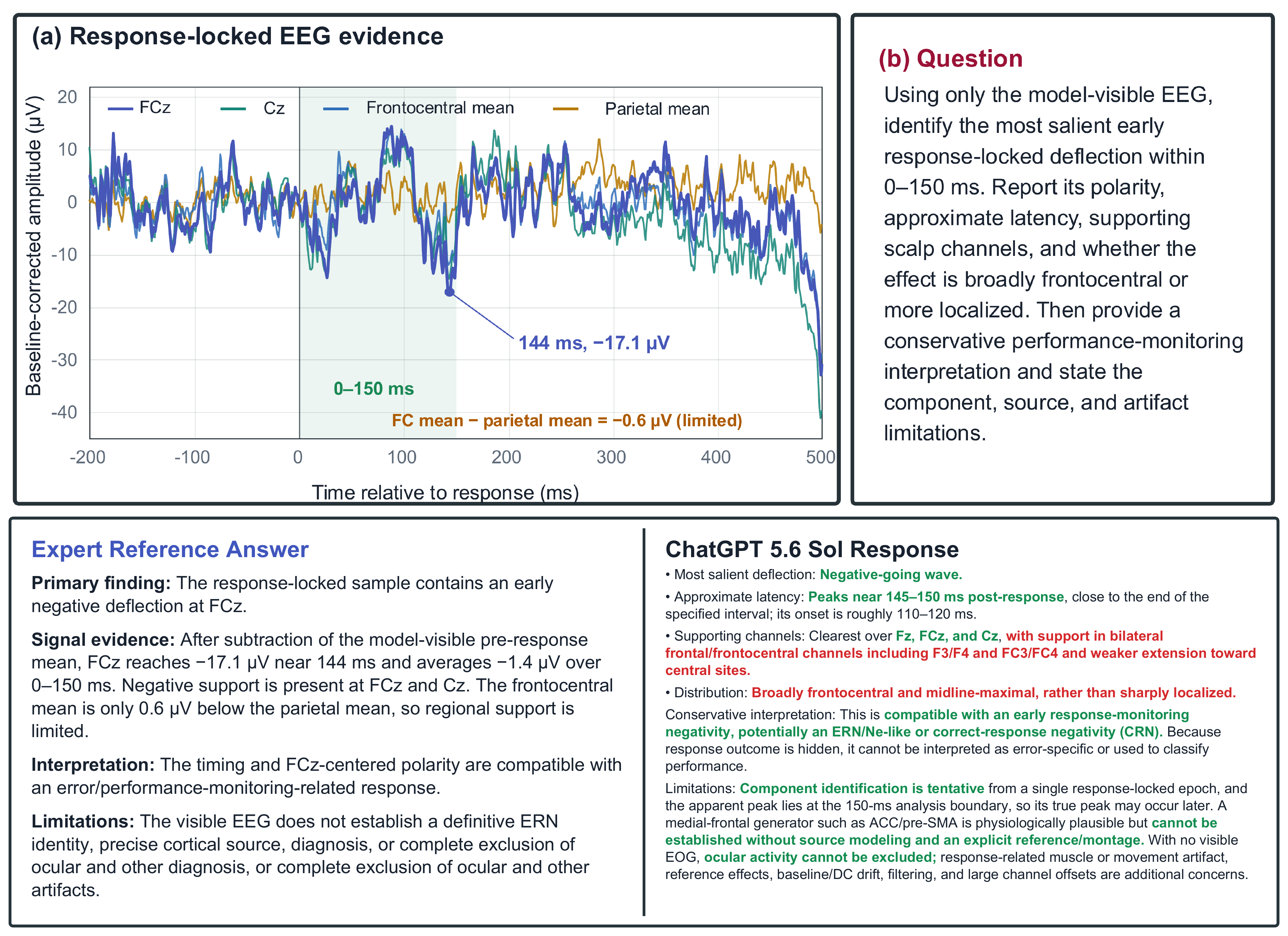}
	\caption{
		Example of rubric-based evaluation for an open-ended EEGProbe question.
		Green text indicates content supported by the expert reference answer,
		while red text indicates incorrect or overstated claims.
	}
	\label{fig:open_ended_evaluation_example}
\end{figure*}

Open-ended tasks require models to interpret the EEG signal and provide a
free-form response. Exact string matching is unsuitable for these tasks,
because semantically equivalent answers may use different wording, while
responses containing correct terminology may still draw unsupported or
incorrect conclusions.
We therefore use a rubric-based evaluation procedure built around
expert-written reference answers. For each question, the reference answer
specifies the expected conclusion, the EEG evidence supporting it, and the
appropriate scope of interpretation. A representative example is shown in
Figure~\ref{fig:open_ended_evaluation_example}.

Each open-ended question is assigned an item-specific weighted rubric covering
the primary finding, supporting signal evidence, neurophysiological
interpretation, and relevant limitations. When rubric components are logically
dependent, score caps are applied. For example, if the primary EEG pattern is
identified incorrectly, downstream claims about its spatial, temporal, or
physiological interpretation cannot receive full credit.
During evaluation, the model response, expert reference answer, and frozen
rubric are provided to multiple independent LLM judges. Each judge assigns
component-level scores and a rationale using the same scoring criteria. The
reference answers and rubrics are used only for evaluation and are never shown
to the model being tested.

The judges assess whether the response:
\begin{enumerate}
	\item identifies the correct primary EEG pattern;
	\item provides signal evidence that is relevant and supported by the input;
	\item gives a neurophysiological interpretation consistent with that evidence;
	\item avoids claims that go beyond what the EEG can support; and
	\item avoids contradictory, unsupported, or irrelevant conclusions.
\end{enumerate}

All models are evaluated using the same judge prompts, reference answers, and
rubrics. Domain experts review the judge outputs and adjudicate cases involving
substantive disagreement or incorrect rubric application. Final open-ended
scores are then normalized to the same 0--100 scale used for the other tasks.

\begin{table*}[t]
	\centering
	\caption{
		Main results on EEGProbe across three evaluation categories and 13 subtasks.
		All values are percentages, and boldface indicates the best result in each column.
		Total is the primary metric.
		Subtasks are abbreviated as follows:
		MI, Motor Imagery Classification;
		ERP, Event-Related Potential Recognition;
		ER, Emotion Recognition;
		SSC, Sleep Stage Classification;
		SD, Seizure Detection;
		EAM, EEG--Audio Matching;
		EIM, EEG--Image Matching;
		ETM, EEG--Text Matching;
		EVM, EEG--Video Matching;
		BRI, Brain-Region Interpretation;
		DQA, Data Quality Assessment;
		TR, Temporal Reasoning;
		and FBA, Frequency-Band Analysis.
	}
	\label{tab:main_results}
	\small
	\setlength{\tabcolsep}{4.0pt}
	\renewcommand{\arraystretch}{1.12}
	
	\begin{tabular*}{\textwidth}{@{\extracolsep{\fill}}lccccccccc@{}}
		\toprule
		& \multicolumn{5}{c}{EEG Signal Classification}
		& \multicolumn{4}{c}{EEG Cross-modal Alignment} \\
		\cmidrule(lr){2-6}\cmidrule(l){7-10}
		Model & MI & ERP & ER & SSC & SD
		& EAM & EIM & ETM & EVM \\
		\midrule
		
		GPT-5.6 Sol
		& \textbf{34.00} & 56.00 & 28.00 & 34.00 & \textbf{64.00}
		& 22.00 & 24.00 & 18.00 & 24.00 \\
		
		Gemini-3.8-Flash$^{\dagger}$
		& 24.00 & \textbf{62.00} & 26.00 & 34.00 & 58.00
		& 24.00 & \textbf{36.00} & 26.00 & 22.00 \\
		
		DeepSeek-V4.1-Flash$^{\dagger}$
		& 26.00 & 40.00 & 28.00 & \textbf{36.00} & 50.00
		& \textbf{28.00} & 24.00 & \textbf{34.00} & 20.00 \\
		
		MiniMax-M3$^{\dagger,\S}$
		& 24.00 & 48.00 & 24.00 & 28.00 & 50.00
		& 20.00 & 30.00 & 20.00 & \textbf{30.00} \\
		
		GPT-5.6 Luna$^{\dagger}$
		& 20.00 & 52.00 & 20.00 & 28.00 & 56.00
		& \textbf{28.00} & \textbf{36.00} & 26.00 & 16.00 \\
		
		GPT-4.1 mini
		& 26.00 & 50.00 & \textbf{40.00} & 26.00 & 54.00
		& \textbf{28.00} & 24.00 & 22.00 & \textbf{30.00} \\
		
		Qwen3.7-Plus$^{\dagger}$
		& 26.00 & 50.00 & 26.00 & 30.00 & 50.00
		& 14.00 & 30.00 & 22.00 & 26.00 \\
		
		Ministral-3-8B
		& 32.00 & 50.00 & 24.00 & 20.00 & 50.00
		& 20.00 & 26.00 & 26.00 & 24.00 \\
		
		Gemma-4-12B
		& 24.00 & 50.00 & 26.00 & 18.00 & 50.00
		& 24.00 & 22.00 & 22.00 & 26.00 \\
		
		Qwen3.5-4B$^{\dagger}$
		& 26.00 & 50.00 & 20.00 & 24.00 & 50.00
		& 22.00 & 24.00 & 22.00 & 18.00 \\
		
		Qwen3-VL-4B-Instruct$^{\ddagger}$
		& 28.00 & 50.00 & 24.00 & 20.00 & 50.00
		& 26.00 & 20.00 & 30.00 & 24.00 \\
		
		\midrule
		Avg. across models
		& 26.36 & 50.73 & 26.00 & 27.09 & 52.91
		& 23.27 & 26.91 & 24.36 & 23.64 \\
		\bottomrule
	\end{tabular*}
	
	\vspace{7pt}
	
	\begin{tabular*}{\textwidth}{@{\extracolsep{\fill}}lcccccc@{}}
		\toprule
		& \multicolumn{4}{c}{EEG Neurophysiological Reasoning}
		& \multicolumn{2}{c}{Overall} \\
		\cmidrule(lr){2-5}\cmidrule(l){6-7}
		Model & BRI & DQA & TR & FBA & Avg. & Total \\
		\midrule
		
		GPT-5.6 Sol
		& 74.93 & 40.00 & \textbf{54.55} & \textbf{80.80}
		& \textbf{42.64} & \textbf{41.63} \\
		
		Gemini-3.8-Flash$^{\dagger}$
		& 69.05 & \textbf{52.00} & 42.42 & 46.70
		& 40.17 & 39.50 \\
		
		DeepSeek-V4.1-Flash$^{\dagger}$
		& 75.20 & 24.00 & 39.39 & 48.10
		& 36.36 & 35.46 \\
		
		MiniMax-M3$^{\dagger,\S}$
		& \textbf{78.85} & 32.00 & 24.24 & 53.15
		& 35.56 & 34.96 \\
		
		GPT-5.6 Luna$^{\dagger}$
		& 57.50 & 40.00 & 21.21 & 50.60
		& 34.72 & 34.61 \\
		
		GPT-4.1 mini
		& 58.31 & 18.00 & 27.27 & 45.40
		& 34.54 & 34.24 \\
		
		Qwen3.7-Plus$^{\dagger}$
		& 66.42 & 30.00 & 24.24 & 49.60
		& 34.17 & 33.77 \\
		
		Ministral-3-8B
		& 39.32 & 16.00 & 33.33 & 47.00
		& 31.36 & 31.14 \\
		
		Gemma-4-12B
		& 44.86 & 24.00 & 24.24 & 36.80
		& 30.15 & 30.00 \\
		
		Qwen3.5-4B$^{\dagger}$
		& 40.34 & 16.00 & 24.24 & 43.20
		& 29.21 & 29.12 \\
		
		Qwen3-VL-4B-Instruct$^{\ddagger}$
		& 20.14 & 18.00 & 27.27 & 28.05
		& 28.11 & 28.30 \\
		
		\midrule
		Avg. across models
		& 56.81 & 28.18 & 31.13 & 48.13
		& 34.27 & 33.88 \\
		\bottomrule
	\end{tabular*}
	
	\vspace{3pt}
	\begin{minipage}{\textwidth}
		\scriptsize
		\textit{Note.} Avg. is the unweighted mean across the 13 subtask scores,
		whereas Total is the instance-weighted aggregate over all 620 benchmark
		instances and serves as the primary metric.
		$^{\dagger}$Provisional BRI/FBA rubric scores.
		$^{\ddagger}$Qwen BRI combines 23 current inputs and 14 earlier lower-resolution inputs.
		$^{\S}$MiniMax DQA follows final-label parsing.
	\end{minipage}
	
\end{table*}

\paragraph{Model Inference}

All models are evaluated on the same frozen benchmark using identical task
inputs, prompts, and candidate options. No model-specific changes are made to
the benchmark or scoring procedure. When an inference interface provides an
optional reasoning or thinking mode, this mode is disabled to maintain a
consistent evaluation setting across models.
For each instance, we record the raw model response, parsed final answer,
execution status, and scoring result. For open-ended questions, we additionally
retain the component-level judge scores, rationales, and expert verification
records. These outputs provide a traceable record of the evaluation process.

\subsection{Comparative Models}

We evaluate 11 contemporary VLMs on EEGProbe:
GPT-5.6 Sol~\citep{openai2026gpt56},
Gemini-3.8-Flash~\citep{google2026gemini38flash},
DeepSeek-V4.1-Flash~\citep{xu2026deepseek},
MiniMax-M3~\citep{lai2026minimax},
GPT-5.6 Luna~\citep{openai2026gpt56},
GPT-4.1 mini~\citep{openai2025gpt41},
Qwen3.7-Plus~\citep{yang2025qwen3},
Ministral-3-8B~\citep{liu2026ministral},
Gemma-4-12B~\citep{team2026gemma},
Qwen3.5-4B~\citep{qwen2026qwen35}, and
Qwen3-VL-4B-Instruct~\citep{qwen2025qwen3}.
The selected models cover different model families, parameter scales, and
deployment settings, including both hosted systems and locally deployable
checkpoints.
For models with downloadable checkpoints, we use the official Hugging Face
weights together with their native tokenizers, multimodal processors, and chat
templates. 
For models with downloadable checkpoints, we deploy the official model
weights on RunPod GPU instances and use vLLM as the inference backend.
Models without local deployment are evaluated through hosted API services,
including the OpenAI API and OpenRouter.
Exact model identifiers are fixed
before evaluation to reduce unintended changes in model versions.
All models are evaluated under the zero-shot protocol described in
Section~\ref{ES}. Each model is tested on the same 620 benchmark instances
covering all 13 subtasks. If an inference request fails to produce a valid
response after the predefined retry procedure, the instance remains in the
evaluation set and is counted as incorrect.

\subsection{Results and Analysis}
\label{sec:results_analysis}

Table~\ref{tab:main_results} reports the performance of the 11 evaluated
VLMs across the 13 EEGProbe subtasks, grouped into three evaluation
categories. We report the score for each subtask, the unweighted mean across
the 13 subtasks (Avg.), and the dataset-level aggregate score over all 620
instances (Total), which serves as the primary metric. 
Because task formats use different scoring procedures, we focus on aggregate
performance, within-subtask comparisons, and broader patterns across the three
evaluation categories.

\paragraph{Overall performance remains limited}
GPT-5.6 Sol achieves the highest Total score of 41.63\%, followed by
Gemini-3.8-Flash at 39.50\%. The mean Total score across all 11 models is
33.88\%, and no model exceeds 50\%. This indicates that current
general-purpose VLMs still have substantial difficulty extracting and using
EEG information across the benchmark. The top two models outperform the
remaining systems, while the models ranked third through seventh fall within
a relatively narrow range of 33.77--35.46\%. Within this group, differences
in overall ranking are small, making their subtask-level performance profiles
more informative than the aggregate scores alone.

\paragraph{Cross-modal EEG alignment is a shared bottleneck}
Performance is consistently low across all four cross-modal subtasks, with
mean scores of 23.27\% on EAM, 26.91\% on EIM, 24.36\% on ETM, and 23.64\%
on EVM. These values remain close to the 25\% chance level of the four-choice
setting, regardless of whether EEG is paired with audio, images, text, or
video. The overall leader, GPT-5.6 Sol, also scores only 18.00--24.00\%
across these subtasks. Moreover, the best individual scores are distributed
across different models rather than concentrated in a single system.
These results suggest that associating EEG responses with external
stimuli remains a common weakness of current VLMs rather than a
modality-specific limitation.

\paragraph{Aggregate performance masks subtask-specific strengths}
No model achieves the best result across all subtasks. GPT-5.6 Sol leads
MI, SD, TR, and FBA. Gemini-3.8-Flash performs best on ERP and DQA and
shares the best EIM score, while DeepSeek-V4.1-Flash leads SSC and ETM and
shares the best EAM result. MiniMax-M3 obtains the highest BRI score and
shares the best EVM result. GPT-5.6 Luna shares the best results on EAM and
EIM, whereas GPT-4.1 mini leads ER and shares the best EAM and EVM scores.
Overall, six models achieve at least one best-in-column result. These results
show that similar aggregate scores can correspond to different capability
profiles across EEGProbe subtasks.

\paragraph{Binary-task plateaus and open-ended dispersion show different performance patterns}
ERP and SD have mean accuracies of 50.73\% and 52.91\%, respectively.
However, six models obtain exactly 50\% on ERP and seven obtain exactly 50\%
on SD. Given the binary response space of these tasks, this concentration
provides little evidence of reliable discrimination and may reflect weak or
collapsed prediction behavior, although accuracy alone cannot determine the
underlying cause. In contrast, the rubric-based BRI and FBA subtasks show
substantially greater cross-model variation, ranging from 20.14--78.85\%
and 28.05--80.80\%, respectively. This wider dispersion suggests that these
open-ended tasks provide additional separation among models when responses
must be supported by region- or frequency-specific EEG evidence. Because
their scores are rubric-based, however, the absolute values should not be
directly compared with exact-match task accuracies. This analysis will be
finalized after the provisional rubric scores in Table~\ref{tab:main_results}
have undergone expert verification.

\paragraph{Model scale alone does not explain the observed ranking}
Among the locally evaluated models with reported parameter counts,
performance is not monotonic with model size. Ministral-3-8B achieves a
Total score of 31.14\%, slightly higher than Gemma-4-12B at 30.00\%, while
the two 4B Qwen models obtain 29.12\% and 28.30\%. This comparison is limited
and does not control for differences in architecture, pretraining data,
multimodal design, or instruction tuning. It nevertheless indicates that
parameter count alone is not a reliable predictor of EEGProbe performance
within this subset of models.

Current VLM performance on EEGProbe remains limited overall and uneven across
subtasks. EEG Cross-modal Alignment is consistently weak, whereas classification
and neurophysiological reasoning show stronger model-specific variation.
These differences are not fully captured by the overall ranking, motivating
a closer analysis of subtask-level strengths and failure patterns.

\section{Detailed Evaluation}

Overall metrics such as Avg. and Total summarize model performance, but they
do not show which properties of EEG current VLMs can actually use. We
therefore examine EEGProbe from three perspectives. First, we characterize
the benchmark in terms of task diversity and its temporal, spectral, and
spatial coverage. Second, we analyze model behavior across categories and
subtasks to identify shared bottlenecks and model-specific strengths. Finally,
we conduct targeted neurophysiological analyses to examine how task-relevant
EEG structure and guidance affect model predictions.

\subsection{Task Diversity and Neurophysiological Coverage}
\label{sec:capability_profiles}

\begin{figure*}[t]
	\centering
	\includegraphics[width=\linewidth]{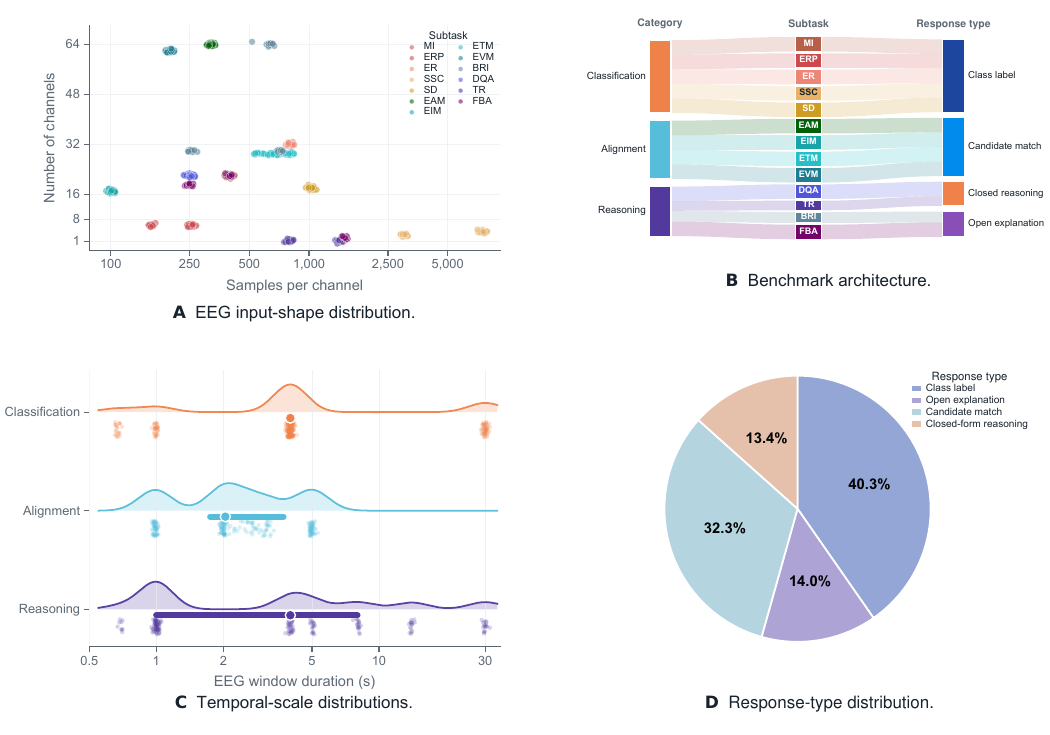}
	\caption{
		\textbf{Benchmark composition and input characteristics of EEGProbe.}
		\textbf{(A)} Distribution of model-visible EEG input shapes.
		\textbf{(B)} Organization of evaluation categories, subtasks, and
		response types.
		\textbf{(C)} EEG window-duration distributions across evaluation
		categories.
		\textbf{(D)} Distribution of task-response types.
	}
	\label{fig:benchmark_structure}
\end{figure*}

EEGProbe contains 13 subtasks grouped into three evaluation categories:
EEG Signal Classification, EEG Cross-modal Alignment, and EEG Neurophysiological
Reasoning. These subtasks cover cognitive, affective, sensorimotor, sleep,
clinical, and signal-quality settings. Figure~\ref{fig:benchmark_structure}
further summarizes the benchmark in terms of input shape, temporal scale,
and response structure.

\begin{figure}[t]
	\centering
	\includegraphics[width=\linewidth]{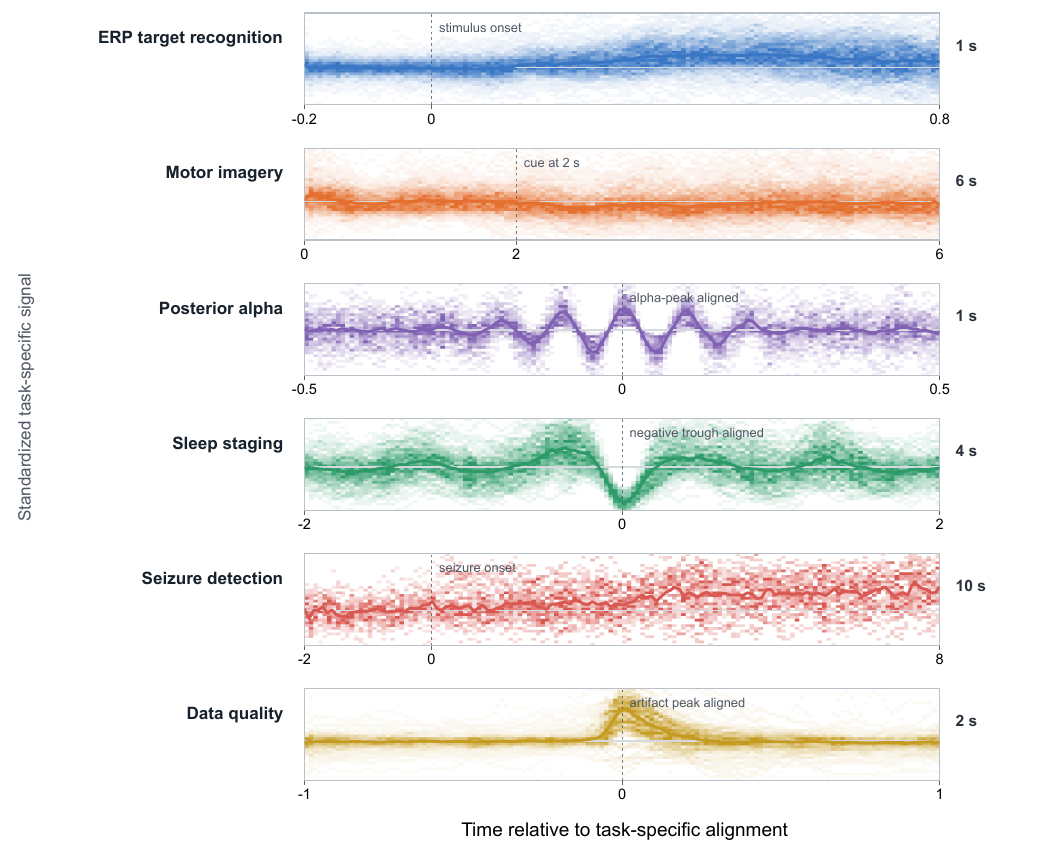}
	\caption{
		Representative temporal patterns from EEGProbe source recordings.
		Signals are aligned to task-specific events or signal landmarks and
		standardized separately within each panel for visualization.
		The figure illustrates differences in temporal morphology and time scale;
		absolute amplitudes should not be compared across datasets.
	}
	\label{fig:temporal_morphology}
\end{figure}

EEGProbe covers a wide range of channel configurations and sequence lengths,
with different subtasks occupying distinct regions of the channel--sample
space. EEG windows range from sub-second segments to approximately 30~s.
The benchmark also contains different response types: 40.3\% of instances
require class-label prediction, 32.3\% candidate matching, 14.0\% open-ended
explanation, and 13.4\% closed-form reasoning. These differences reflect the
heterogeneity of both the EEG inputs and the tasks constructed from them.
We next examine whether this task diversity is also reflected in the
underlying signal characteristics. We consider temporal, spectral, and
spatial properties of representative EEG recordings, beginning with temporal
morphology.
Figure~\ref{fig:temporal_morphology} shows representative temporal patterns
from several EEGProbe source regimes. The examples include event-locked
responses, sustained oscillatory activity, slow-wave dynamics, evolving
seizure activity, and physiological artifacts, which occur over substantially
different time scales.

The signals are aligned to task-specific events or landmarks, such as
stimulus onset, motor-imagery cue, oscillatory peak, seizure onset, or
artifact peak. As the recordings come from different datasets, the figure
is intended to compare temporal patterns rather than absolute amplitudes.

The temporal patterns in Figure~\ref{fig:temporal_morphology} reveal that
EEGProbe spans fundamentally different forms of electrophysiological
organization. ERP responses are dominated by transient, event-locked activity
following stimulus onset, whereas posterior alpha is expressed through
recurring oscillatory cycles. Sleep-related activity and seizure recordings
extend over longer temporal contexts, while artifact-related signals exhibit
distinct localized temporal structure. The benchmark therefore covers a
continuum from subsecond evoked responses to sustained rhythms and multisecond
physiological or pathological dynamics.
This temporal diversity is important because the relevant evidence is not
expressed at a single time scale. Depending on the subtask, a model may need
to detect a localized event, recognize a recurring oscillatory pattern, or
integrate information over an evolving temporal trajectory.
EEGProbe therefore tests whether VLMs can handle EEG patterns that differ
substantially in their temporal scale and organization.

\begin{figure}[t]
	\centering
	\includegraphics[width=\linewidth]{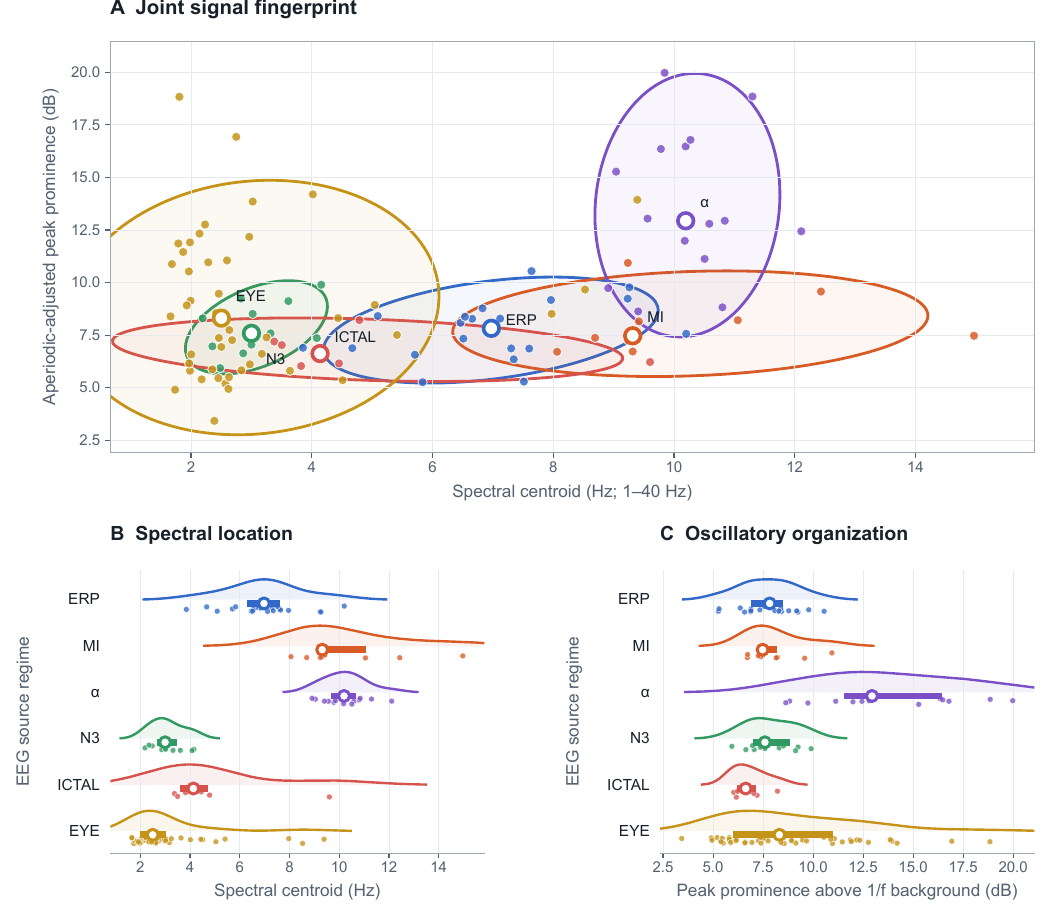}
	\caption{
		Spectral characteristics of representative EEG regimes in EEGProbe.
		\textbf{(A)} Joint distribution of spectral centroid and
		aperiodic-adjusted peak prominence.
		\textbf{(B)} Subject-level distributions of spectral centroid.
		\textbf{(C)} Subject-level distributions of peak prominence above the
		fitted $1/f$ background.
		Each point represents a subject-level median; open circles indicate
		group medians, and the ellipses in (A) denote descriptive 80\%
		covariance regions. These descriptors are used only for benchmark
		characterization and are not provided as model inputs.
	}
	\label{fig:spectral_coverage}
\end{figure}

We further examine whether this diversity extends to the frequency domain.
Figure~\ref{fig:spectral_coverage} characterizes representative source
recordings using two complementary descriptors: spectral centroid, which
captures the frequency-weighted location of power within 1--40~Hz, and
aperiodic-adjusted peak prominence, which measures the strength of the
dominant oscillatory peak relative to the broadband aperiodic background.
These descriptors are used only for benchmark characterization and are not
provided to the evaluated models.

The resulting distributions show that EEGProbe spans several distinct
spectral regimes rather than a narrowly concentrated signal space.
N3 sleep and ocular activity occupy predominantly low-frequency regions,
whereas motor-imagery and posterior-alpha recordings shift toward higher
spectral centroids. Posterior alpha additionally exhibits pronounced
narrowband organization, while motor imagery shows a broader spectral
profile. ERP and ictal recordings occupy intermediate regions, producing a
spectral landscape with both overlap and clear regime-specific structure.
The subject-level distributions in
Figure~\ref{fig:spectral_coverage}(B) and Figure~\ref{fig:spectral_coverage}(C) further show that these differences
are consistently observed across subjects and are not limited to a small
number of epochs.
Importantly, the benchmark varies along two complementary spectral axes:
where signal power is concentrated and how strongly that power is organized
into oscillatory structure. As a result, EEGProbe includes slow-wave,
event-related, sensorimotor, posterior-rhythmic, pathological, and artifactual
regimes within a common evaluation framework.

Together, these analyses show that EEGProbe covers a wide range of EEG
patterns, from short event-related responses to sustained oscillatory and
multisecond dynamics, with clear differences in spectral organization. This diversity creates a
more demanding test of whether general-purpose VLMs can extract useful
evidence across heterogeneous neural-signal structures rather than succeed
only within a narrow class of EEG patterns.

\subsection{What Do Current Models Understand from EEG?}
\label{5.2}

\paragraph{From aggregate ranking to EEG capability profiles}
Although Avg. and Total produce the same overall model ranking, this agreement
masks substantial differences in how that performance is distributed across
EEG capabilities. GPT-5.6 Sol and Gemini-3.8-Flash remain the two strongest
models overall, with Total scores of 41.63\% and 39.50\%, respectively, while
several models below them occupy a much narrower performance range. We
therefore move beyond the aggregate leaderboard and examine how each model
performs across evaluation categories and individual subtasks.

At the category level, GPT-5.6 Sol leads EEG Signal Classification and
Neurophysiological Reasoning, with mean scores of 43.20\% and 62.57\%,
respectively, whereas Gemini-3.8-Flash performs better on cross-modal
alignment (27.00\% versus 22.00\%). Models with nearly identical overall
scores can exhibit even larger differences in their internal capability
profiles. MiniMax-M3 and GPT-4.1 mini, for example, differ by only
0.72 percentage points in Total, yet MiniMax-M3 is substantially stronger in
reasoning (47.06\% versus 37.25\%), while GPT-4.1 mini performs better in
classification (39.20\% versus 34.80\%) and slightly better in alignment
(26.00\% versus 25.00\%). Category averages are computed as the unweighted
mean of the corresponding subtask scores.

\begin{figure*}[t]
	\centering
	\includegraphics[width=\textwidth]{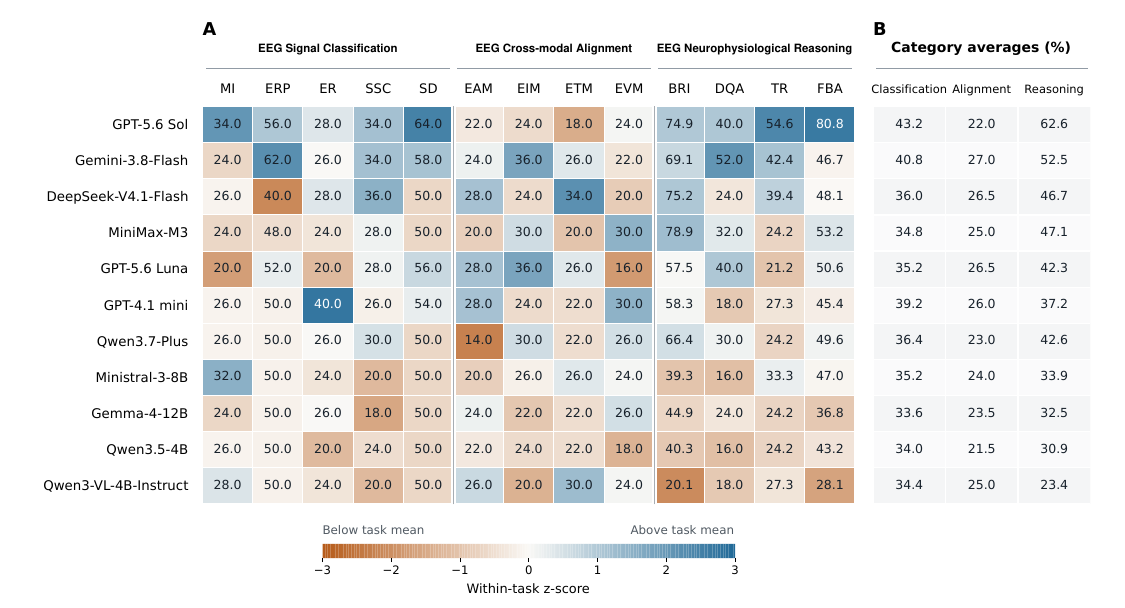}
	\caption{
		Model capability profiles across EEGProbe. Models are ordered by descending
		Total score.
		\textbf{(A)} Scores across the 13 subtasks, grouped into EEG Signal
		Classification, EEG Cross-modal Alignment, and EEG Neurophysiological
		Reasoning. Cell values show raw scores (\%), while color represents the
		within-subtask $z$-score across the 11 models.
		\textbf{(B)} Unweighted mean score within each evaluation category.
	}
	\label{fig:capability_heatmap}
\end{figure*}

Figure~\ref{fig:capability_heatmap} further shows how these differences are
distributed across evaluation categories and subtasks. Panel B summarizes the
category-level profiles. GPT-5.6 Sol achieves the highest mean scores in EEG
Signal Classification and EEG Neurophysiological Reasoning, whereas
Gemini-3.8-Flash performs best in EEG Cross-modal Alignment.
Panel A shows where these differences arise at the subtask level. GPT-5.6 Sol
is strongest on MI, SD, TR, and FBA, while Gemini-3.8-Flash leads ERP and DQA
and shares the highest EIM score. Other models show more localized strengths:
DeepSeek-V4.1-Flash performs best on SSC and ETM, MiniMax-M3 achieves the
highest BRI score, and GPT-4.1 mini leads ER. The resulting profiles differ
substantially even among models with similar overall scores.
Each cell in Panel A reports the raw benchmark score, while the color represents
the within-subtask $z$-score across the 11 models. The color scale is used only
to show relative performance within each subtask.

The same pattern appears among models with similar category averages.
DeepSeek-V4.1-Flash and GPT-5.6 Luna, for instance, both obtain an alignment
average of 26.50\%, yet their subtask behavior differs markedly:
DeepSeek-V4.1-Flash scores higher on EEG--text matching
(34.00\% versus 26.00\%), whereas GPT-5.6 Luna performs better on
EEG--image matching (36.00\% versus 24.00\%). These results show that EEG
capability is not well described by a single scalar score. Aggregate metrics
capture overall performance, but category- and subtask-level profiles reveal
which forms of EEG evidence each model can use more effectively.

\begin{figure}[t]
	\centering
	\includegraphics[width=\linewidth]{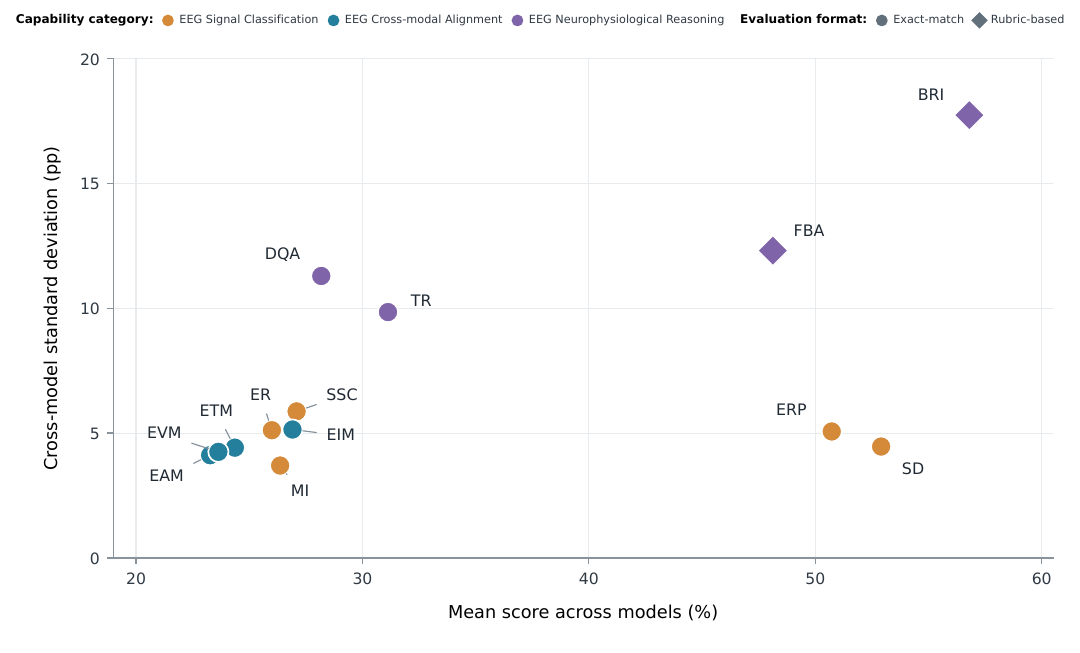}
	\caption{
		Subtask-level diagnostic map across the 11 evaluated models.
		Each subtask is positioned by its mean score and cross-model population
		standard deviation. Colors denote the three evaluation categories, while
		marker shapes distinguish exact-match from rubric-based evaluation.
		Higher cross-model dispersion indicates stronger empirical separation
		among models.
	}
	\label{fig:task_diagnostic_map}
\end{figure}

\paragraph{Shared bottlenecks and capability-discriminative subtasks}
Figure~\ref{fig:task_diagnostic_map} provides a task-centric view of EEGProbe.
Each subtask is characterized by its mean score across the 11 evaluated models
and the corresponding cross-model standard deviation. The former reflects the
performance level reached by current models, while the latter indicates how
strongly the subtask separates their capabilities.

The four EEG Cross-modal Alignment subtasks form the clearest low-performance
cluster. Their mean scores are 23.27\% for EAM, 26.91\% for EIM, 24.36\% for
ETM, and 23.64\% for EVM, with cross-model standard deviations of only
4.11--5.14 percentage points. Despite involving different external modalities,
all four remain close to the 25\% chance level of their four-choice protocols
and show similarly limited variation across models. This consistency points to
a shared bottleneck in using EEG evidence for stimulus-level cross-modal
matching rather than a weakness specific to any one modality. MI, ER, and SSC
also occupy the low-mean region, indicating that limited performance extends
to several direct EEG classification subtasks.

EEG Neurophysiological Reasoning shows a markedly different pattern. BRI reaches
a mean score of 56.81\% with a cross-model standard deviation of 17.74
percentage points, while FBA achieves 48.13\% with a standard deviation of
12.31 points. These two subtasks provide the strongest empirical separation
among the evaluated models. Importantly, this pattern is not confined to
rubric-based evaluation: DQA and TR, both exact-match tasks, also exhibit
substantial cross-model variation, with standard deviations of 11.30 and
9.85 percentage points, respectively. This suggests that neurophysiological
reasoning exposes larger differences in how effectively current models use
task-relevant EEG evidence. By contrast, ERP and SD
show higher mean scores but relatively low dispersion, consistent with the
binary-task plateaus observed in the main results.

Figures~\ref{fig:capability_heatmap} and
Figures~\ref{fig:task_diagnostic_map} reveal two complementary properties of current
EEG capability. First, models with similar aggregate scores can rely on very
different combinations of subtask-level strengths. Second, the subtasks
themselves differ substantially in diagnostic value: EEG Cross-modal Alignment
reveals a common failure mode, whereas EEG Neurophysiological Reasoning produces
much stronger separation among models. Current EEG capability is therefore
both model-dependent and task-dependent, rather than reducible to a single
overall score.

The next section examines whether explicitly directing models toward
task-relevant neurophysiological evidence can alter these prediction patterns.

\subsection{Can Neurophysiological Guidance Improve EEG Predictions?}

\begin{figure}[t]
	\centering
	\includegraphics[width=\linewidth]{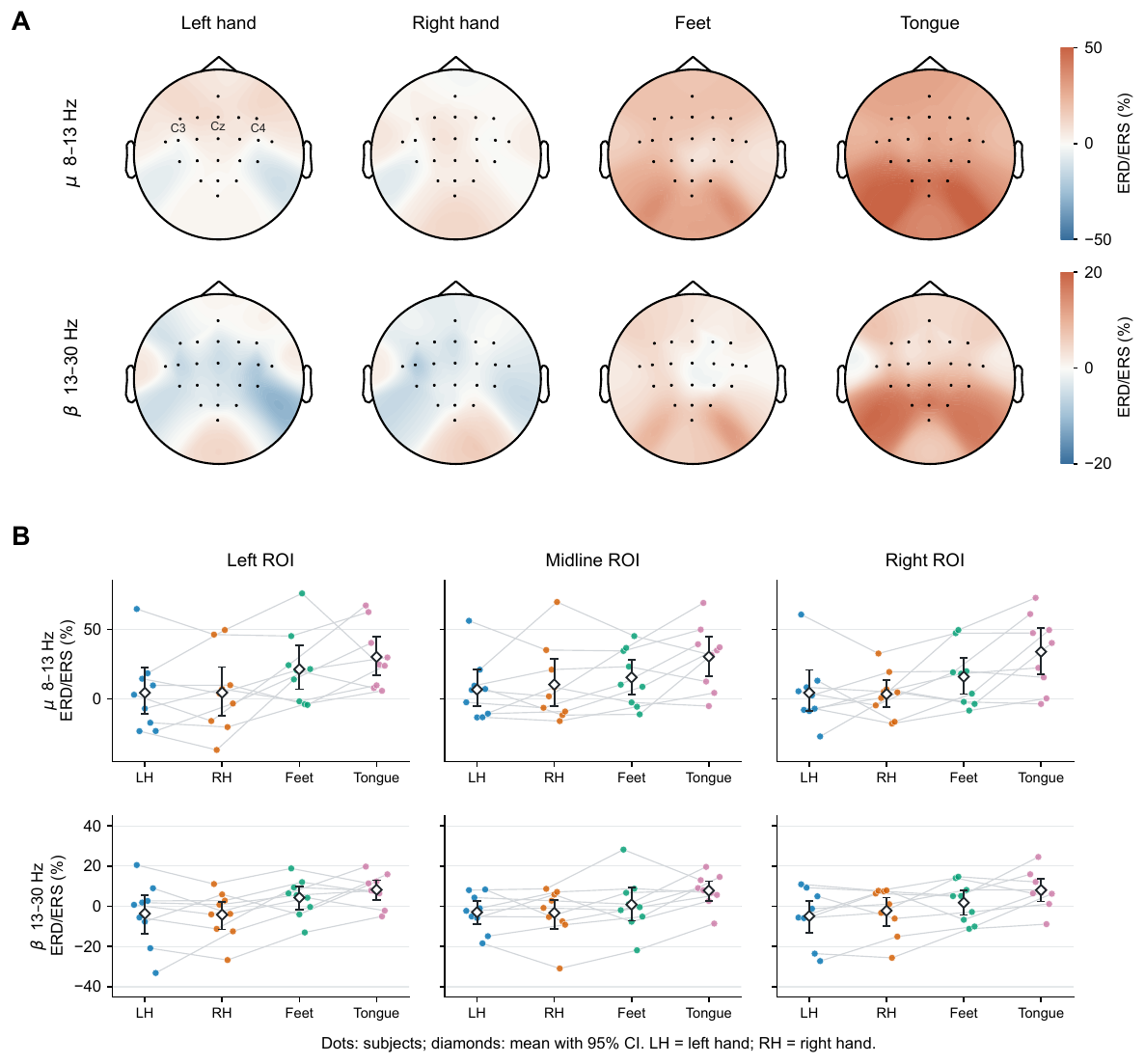}
	\caption{
		Neurophysiological patterns associated with motor-imagery classes.
		\textbf{(A)} Group-level scalp distributions of $\mu$-band
		(8--13 Hz) and $\beta$-band (13--30 Hz) ERD/ERS for left-hand,
		right-hand, feet, and tongue imagery.
		Negative values indicate power suppression relative to the pre-cue
		baseline, whereas positive values indicate power increases.
		\textbf{(B)} Subject-level ERD/ERS across left
		(FC3--C3--CP3), midline (FCz--Cz--CPz), and right
		(FC4--C4--CP4) sensorimotor ROIs.
		Dots denote individual subjects, and diamonds indicate group means with
		95\% confidence intervals.
		The analysis uses 2,283 artifact-screened, benchmark-trial-disjoint
		trials from nine subjects in BCI Competition IV-2a.
	}
	\label{fig:mi_physiology}
\end{figure}

The preceding analysis shows that current VLMs have uneven EEG capabilities:
EEG Cross-modal Alignment remains a shared bottleneck, while reasoning tasks
produce much stronger model separation. We next ask whether these limitations
reflect an inability to use task-relevant EEG structure that is already present
in the signal.
We use fixed inference-time guidance derived from independently characterized
neurophysiological structure, without updating model parameters. For
representative EEG Signal Classification and EEG Cross-modal Alignment tasks,
we first analyze the EEG evidence relevant to the target discrimination and
identify recurrent model failure patterns on a diagnosis subset. The resulting
guidance is then frozen and applied unchanged to held-out questions, with the
EEG input, question wording, answer choices, and option order kept fixed.
This design allows us to test whether explicitly directing the model toward
task-relevant neural evidence changes its prediction behavior.

\paragraph{Neurophysiology-guided motor-imagery classification}

We first study motor imagery as a representative signal-classification task.
Using 2,283 artifact-screened, benchmark-trial-disjoint trials from nine
subjects in BCI Competition IV-2a, we characterize class-dependent
spatial--spectral structure in the $\mu$ (8--13 Hz) and $\beta$ (13--30 Hz)
bands. As shown in Figure~\ref{fig:mi_physiology}, the four imagery classes
exhibit different patterns across left, midline, and right sensorimotor
regions. Several contrasts are also consistent across subjects: relative to
tongue imagery, right-hand imagery shows stronger left-ROI $\beta$
suppression in all nine subjects, while left-hand imagery shows stronger
left-ROI $\mu$ suppression in seven of nine subjects. Feet imagery similarly
shows stronger right-ROI $\mu$ suppression than tongue imagery in seven of
nine subjects. These results indicate that MI classes are associated with
joint spatial--spectral patterns that extend beyond a simple hemispheric rule.

We next analyzed GPT-5.6 Sol on a 30-question diagnosis subset. Among the
21 incorrect predictions, 14 mapped feet or tongue trials to one of the two
hand classes, while six reflected confusion between left- and right-hand
imagery. Based on these failure patterns and the independently characterized
EEG structure, we constructed a single task-level guidance prompt. The prompt
directs the model to jointly examine $\mu$- and $\beta$-band structure across
left, midline, and right sensorimotor regions, integrate evidence across
neighboring channels, and avoid assigning a class from a single regional cue.

The benchmark provides only a 4-s post-cue EEG window and does not include a
pre-cue baseline. The guidance therefore does not require the model to compute
baseline-relative ERD/ERS directly. Instead, it focuses on within-window
spatial--spectral differences that remain available in the model-visible
signal. The guidance was frozen after the 30-question diagnosis subset and
applied unchanged to the remaining 20 held-out MI questions.

The baseline obtained 8/20 correct predictions (40.0\%) on the held-out set,
but all 20 predictions were assigned to either the left-hand or right-hand
class. Eight of the ten hand trials were classified correctly, whereas none
of the ten feet or tongue trials were correctly identified. The model
therefore achieved its baseline accuracy almost entirely through the two hand
classes, despite the class-dependent spatial--spectral structure observed in
the independent physiological analysis.

Guidance substantially changed this prediction pattern. Hand-class predictions
fell from 20/20 to 8/20, all four MI labels were used, and feet/tongue-to-hand
errors decreased from 10 to 4. Accuracy on the ten non-hand trials increased
from 0/10 to 5/10. These gains were offset by additional errors on hand trials,
leaving the overall accuracy unchanged at 8/20 (40.0\%).

The intervention therefore did not improve aggregate MI accuracy, but it
changed the class distribution of the predictions in a direction consistent
with the independently observed sensorimotor structure. The baseline largely
ignored distinctions involving feet and tongue imagery, whereas the guided
model began to separate these classes from hand imagery. This result suggests
that task-relevant EEG structure can be present in the signal without being
used reliably by the model, and that explicit neurophysiological guidance can
change which distinctions influence its predictions.

\paragraph{Neurophysiology-guided EEG--image matching}

\begin{figure*}[t]
	\centering
	\includegraphics[width=\textwidth]{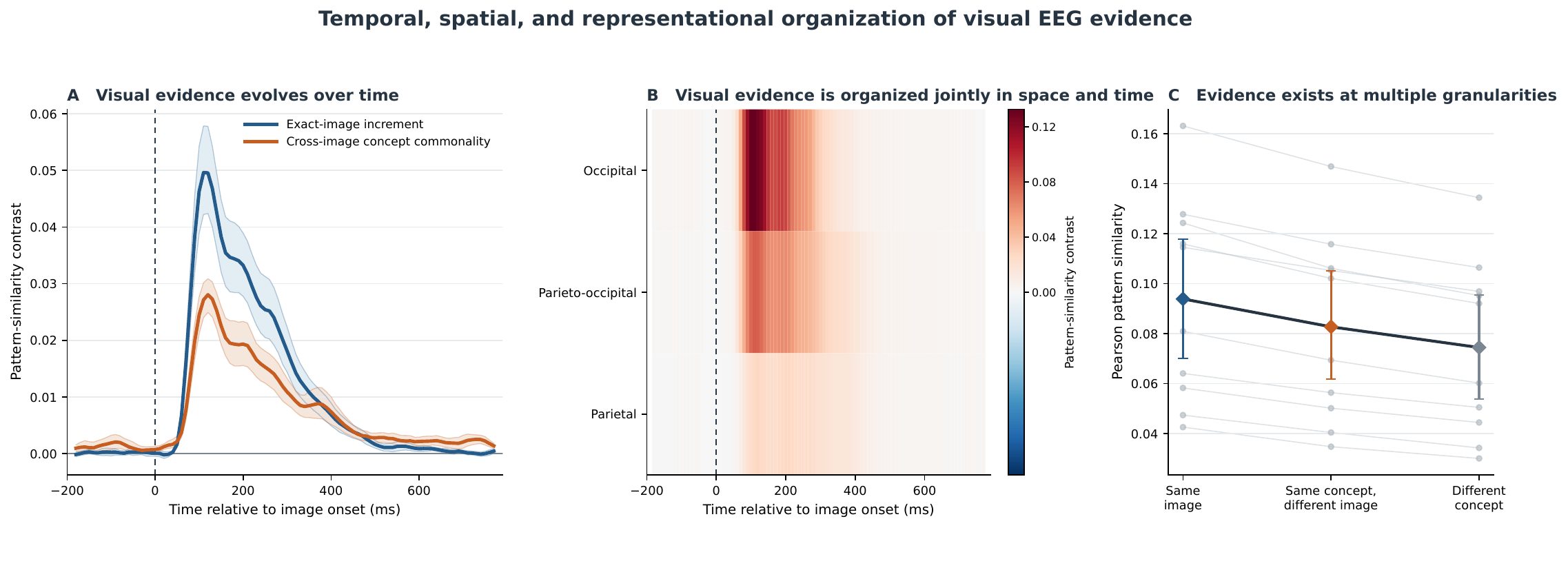}
	\caption{
		Temporal, spatial, and representational organization of visual EEG
		evidence in benchmark-disjoint THINGS-EEG2 training recordings.
		\textbf{(A)} Stimulus-locked temporal profiles of image-specific and
		concept-level pattern-similarity contrasts.
		\textbf{(B)} Spatial--temporal distribution of visual evidence across
		posterior channel groups.
		\textbf{(C)} Representational similarity for repeated images,
		different images from the same concept, and images from different
		concepts.
		All ten participants exhibit the ordering
		same image $>$ same concept but different image $>$ different concept.
	}
	\label{fig:eim_physiology}
\end{figure*}

We next examine EEG--image matching as a representative cross-modal
alignment task. Using benchmark-disjoint THINGS-EEG2 training recordings,
we first characterize the visual information available in the EEG signal.
Figure~\ref{fig:eim_physiology} shows clear temporal, spatial, and
representational structure.

Figure~\ref{fig:eim_physiology}(A) shows that both image-specific and concept-level pattern similarity
emerge after stimulus onset, with the image-specific contrast reaching a
larger peak. The EEG therefore contains information related both to
visual concepts and to individual stimuli. Figure~\ref{fig:eim_physiology}(B) shows that this evidence
is strongest over posterior regions, particularly occipital channels, and changes
over post-stimulus time. Figure~\ref{fig:eim_physiology}(C) further shows a consistent representational
ordering across all ten participants:
same image $>$ same concept but different image $>$ different concept.
Thus, visual EEG contains information at more than one level of
specificity, from shared concept-level structure to finer image-specific
structure.
This distinction is important for EEG--image matching. Evidence that supports
a general visual concept may still be insufficient to identify one image
among several semantically related candidates. EIM therefore depends
on whether the model can use the temporal and posterior EEG structure at the
level of specificity required by the candidate set.

We then analyzed GPT-5.6 Sol on a frozen 30-question diagnosis subset.
Among the 22 incorrect responses, 11 cited EEG evidence that did not
discriminate the selected image from other candidates. Six responses
interpreted generic signal variation as specific semantic content, while
five favored visually salient candidates without corresponding discriminative
EEG support. These errors point to a recurring problem: the specificity of the final image choice often exceeded the specificity supported by the EEG evidence.

Based on these errors and the independently characterized visual EEG
structure, we constructed a single fixed guidance prompt. The guidance first
asks the model to identify how the candidate images differ and then assess
whether the EEG contains evidence at the same level of specificity. Following
the structure shown in Figure~\ref{fig:eim_physiology}, it emphasizes
stimulus-locked temporal patterns, distributed posterior activity, and the
distinction between concept-level and image-specific evidence. EEG evidence
is treated as discriminative only when it helps separate one candidate from
the alternatives. The guidance was frozen after analysis of the diagnosis
subset and applied unchanged to the remaining 20 held-out EIM questions.
The EEG input, candidate images, candidate order, and answer format were kept
fixed.

On the held-out subset, baseline accuracy was 4/20 (20.0\%), while the guided
condition achieved 8/20 (40.0\%), as shown in
Table~\ref{tab:eim_guided_results}. Five previously incorrect responses became
correct, one previously correct response became incorrect, three remained
correct, and 11 remained incorrect. Predictions continued to span all four
answer positions, so the improvement was not associated with concentration on
a single option. In this held-out subset, guidance that matched the
specificity of the EEG evidence to the required image decision increased EIM
accuracy by 20 percentage points.

\begin{table}[t]
	\centering
	\caption{
		Held-out performance of GPT-5.6 Sol on EEG--image matching before and
		after neurophysiological guidance.
	}
	\label{tab:eim_guided_results}
	\begin{tabular}{lcc}
		\toprule
		Method & Correct & Accuracy \\
		\midrule
		Baseline & 4/20 & 20.0\% \\
		Baseline + guidance & 8/20 & 40.0\% \\
		\bottomrule
	\end{tabular}
\end{table}

\paragraph{Source-specific failure patterns in data-quality assessment}
We finally examine DQA as a diagnostic case of EEG Neurophysiological Reasoning.
Across 50 one-second, 22-channel EEG segments, GPT-5.6 Sol achieved
20/50 correct predictions (40.0\%), but performance varied substantially
across artifact types. Recall was highest for ocular artifacts (8/9) and
electrode disturbances (7/8), followed by muscle (3/8) and chewing artifacts
(2/8). No shivering (0/8) or background cases (0/9) were correctly identified.

A strong electrode-artifact bias was also evident: 28 of the 50 segments were
predicted as electrode disturbances, including all eight shivering cases,
yielding only 25.0\% precision for this class. The model therefore appears
better at detecting conspicuous abnormality than at identifying its physical
source.

This distinction is important because artifact attribution requires different
forms of evidence. Ocular activity is often characterized by its waveform
morphology and frontal distribution, whereas muscle-related contamination
depends more strongly on spectral breadth, burst structure, and spatial
extent. Chewing and shivering introduce additional temporal regularities that
can overlap with other forms of contamination. Reliable source attribution
therefore requires joint use of temporal, spectral, and cross-channel
evidence rather than detection of sharp or irregular activity alone.

The failure on all nine background segments further shows that DQA involves
more than artifact detection: the model must also recognize when no dominant
artifact source is present. Taken together, these results reveal a clear
asymmetry in current EEG reasoning. Salient abnormalities such as ocular
artifacts can often be recognized, while source-level attribution becomes
much less reliable when multiple signal dimensions must be integrated.

Across the three analyses, a consistent picture emerges. In MI, guidance
changed a collapsed hand-dominated prediction pattern without improving
overall accuracy. In EIM, guidance improved held-out accuracy from 20.0\% to
40.0\%. DQA, meanwhile, shows that source attribution remains difficult even
when abnormal signal structure is detectable. These findings suggest that
current VLM limitations are not explained only by the absence of useful EEG
information. In several settings, task-relevant structure is present in the
signal but is not used reliably by the model. Making that structure explicit
can alter inference behavior, although the effect depends strongly on the
task and the specificity of the available evidence.

\section{Conclusion}

We introduced EEGProbe, a signal-grounded benchmark for evaluating
whether general-purpose VLMs can extract and use information from minimally
processed EEG. EEGProbe contains 620 questions across 13 subtasks in three
evaluation categories: EEG Signal Classification, EEG Cross-modal Alignment,
and EEG Neurophysiological Reasoning. Signal-necessity testing and expert
verification are used to ensure that successful answers depend on EEG evidence
rather than textual shortcuts.
Our results reveal a central limitation of current VLMs: \textit{informative EEG structure may be present in the signal, yet models often fail to use it reliably}. The strongest model reaches only 41.63\% overall, and cross-modal
alignment remains close to chance.
Current VLMs show several clear weaknesses: unstable class discrimination, poor use of fine-grained EEG evidence, and limited source-level neurophysiological reasoning. These failures suggest that useful EEG structure is still not extracted or used reliably.
Neurophysiological analyses further show
that explicitly directing models toward task-relevant EEG structure can alter
their predictions and, in some cases, improve performance. This highlights an important distinction between what a model can extract from EEG on its own and what it can achieve when task-specific neurophysiological knowledge is supplied at inference time.
These findings point to a clear direction for future EEG-capable multimodal
models: they should require less task-specific guidance and become better able
to extract task-relevant electrophysiological structure,
align it with the required decision, and use it consistently during inference. 
EEGProbe provides a testbed for measuring progress toward this goal.

\section*{CRediT authorship contribution statement}

\textbf{Haozhe Zhu:} Conceptualization, Investigation, Data curation, Project administration,
Writing -- original draft, Writing -- review \& editing.
\textbf{Dan Yuan:} Conceptualization, Supervision, Writing -- original draft,
Writing -- review \& editing.
\textbf{Chenghao Gu:} Investigation.
\textbf{Feike Xu:} Investigation.
\textbf{Yue Wen:} Conceptualization, Writing -- review \& editing.
\textbf{Shin'ichi Konomi:} Supervision, Funding acquisition,
Writing -- review \& editing.

\section*{Declaration of competing interest}

The authors declare that they have no known competing financial interests or
personal relationships that could have appeared to influence the work reported
in this paper.

\section*{Acknowledgment}

This work was supported by JSPS KAKENHI Grant Numbers JP23H03507 and JP23K02469, and JST SPRING, Grant Number JPMJSP2136.

\section*{Data availability}

The EEGProbe benchmark data and evaluation resources will be made publicly
available upon publication.

\bibliographystyle{elsarticle-harv}
\bibliography{eegbib}

@inproceedings{radford2021learning,
  title={Learning transferable visual models from natural language supervision},
  author={Radford, Alec and Kim, Jong Wook and Hallacy, Chris and Ramesh, Aditya and Goh, Gabriel and Agarwal, Sandhini and Sastry, Girish and Askell, Amanda and Mishkin, Pamela and Clark, Jack and others},
  booktitle={International conference on machine learning},
  pages={8748--8763},
  year={2021},
  organization={PmLR}
}

@article{al2021deep,
  title={Deep learning for motor imagery EEG-based classification: A review},
  author={Al-Saegh, Ali and Dawwd, Shefa A and Abdul-Jabbar, Jassim M},
  journal={Biomedical Signal Processing and Control},
  volume={63},
  pages={102172},
  year={2021},
  publisher={Elsevier}
}

@article{padfield2019eeg,
  title={EEG-based brain-computer interfaces using motor-imagery: Techniques and challenges},
  author={Padfield, Natasha and Zabalza, Jaime and Zhao, Huimin and Masero, Valentin and Ren, Jinchang},
  journal={Sensors},
  volume={19},
  number={6},
  pages={1423},
  year={2019},
  publisher={MDPI}
}

@article{jenke2014feature,
  title={Feature extraction and selection for emotion recognition from EEG},
  author={Jenke, Robert and Peer, Angelika and Buss, Martin},
  journal={IEEE Transactions on Affective computing},
  volume={5},
  number={3},
  pages={327--339},
  year={2014},
  publisher={IEEE}
}

@article{jafari2023emotion,
  title={Emotion recognition in EEG signals using deep learning methods: A review},
  author={Jafari, Mahboobeh and Shoeibi, Afshin and Khodatars, Marjane and Bagherzadeh, Sara and Shalbaf, Ahmad and Garc{\'\i}a, David L{\'o}pez and Gorriz, Juan M and Acharya, U Rajendra},
  journal={Computers in biology and medicine},
  volume={165},
  pages={107450},
  year={2023},
  publisher={Elsevier}
}

@article{aboalayon2016sleep,
  title={Sleep stage classification using EEG signal analysis: a comprehensive survey and new investigation},
  author={Aboalayon, Khald Ali I and Faezipour, Miad and Almuhammadi, Wafaa S and Moslehpour, Saeid},
  journal={Entropy},
  volume={18},
  number={9},
  pages={272},
  year={2016},
  publisher={MDPI}
}

@article{phan2022automatic,
  title={Automatic sleep staging of EEG signals: recent development, challenges, and future directions},
  author={Phan, Huy and Mikkelsen, Kaare},
  journal={Physiological measurement},
  volume={43},
  number={4},
  pages={04TR01},
  year={2022},
  publisher={IOP Publishing}
}

@article{boonyakitanont2020review,
  title={A review of feature extraction and performance evaluation in epileptic seizure detection using EEG},
  author={Boonyakitanont, Poomipat and Lek-Uthai, Apiwat and Chomtho, Krisnachai and Songsiri, Jitkomut},
  journal={Biomedical Signal Processing and Control},
  volume={57},
  pages={101702},
  year={2020},
  publisher={Elsevier}
}

@article{ein2023eeg,
  title={EEG seizure detection: concepts, techniques, challenges, and future trends},
  author={Ein Shoka, Athar A and Dessouky, Mohamed M and El-Sayed, Ayman and Hemdan, Ezz El-Din},
  journal={Multimedia tools and applications},
  volume={82},
  number={27},
  pages={42021--42051},
  year={2023},
  publisher={Springer}
}

@article{light2010electroencephalography,
  title={Electroencephalography (EEG) and event-related potentials (ERPs) with human participants},
  author={Light, Gregory A and Williams, Lisa E and Minow, Falk and Sprock, Joyce and Rissling, Anthony and Sharp, Richard and Swerdlow, Neal R and Braff, David L},
  journal={Current protocols in neuroscience},
  volume={52},
  number={1},
  pages={6--25},
  year={2010},
  publisher={Wiley Online Library}
}

@article{fietz2025empathy,
  title={Empathy and event related potentials before and after EEG based neurofeedback training in autistic adolescents},
  author={Fietz, Julia and Auer, Gabriel and Plener, Paul and Poustka, Luise and Konicar, Lilian},
  journal={Scientific Reports},
  volume={15},
  number={1},
  pages={30824},
  year={2025},
  publisher={Nature Publishing Group UK London}
}

@article{zhao2025tcanet,
  title={TCANet: a temporal convolutional attention network for motor imagery EEG decoding},
  author={Zhao, Wei and Lu, Haodong and Zhang, Baocan and Zheng, Xinwang and Wang, Wenfeng and Zhou, Haifeng},
  journal={Cognitive Neurodynamics},
  volume={19},
  number={1},
  pages={91},
  year={2025},
  publisher={Springer}
}

@article{zhang2024vision,
  title={Vision-language models for vision tasks: A survey},
  author={Zhang, Jingyi and Huang, Jiaxing and Jin, Sheng and Lu, Shijian},
  journal={IEEE transactions on pattern analysis and machine intelligence},
  volume={46},
  number={8},
  pages={5625--5644},
  year={2024},
  publisher={IEEE}
}

@article{chen2025eeg,
  title={EEG-based affective brain--computer interfaces: Recent advancements and future challenges},
  author={Chen, Yuxin and Peng, Yong and Tang, Jiajia and Camilleri, Tracey and Camilleri, Kenneth and Kong, Wanzeng and Cichocki, Andrzej},
  journal={Journal of Neural Engineering},
  volume={22},
  number={3},
  pages={031004},
  year={2025},
  publisher={IOP Publishing}
}

@article{liu2025recent,
  title={Recent applications of EEG-based brain-computer-interface in the medical field},
  author={Liu, Xiu-Yun and Wang, Wen-Long and Liu, Miao and Chen, Ming-Yi and Pereira, T{\^a}nia and Doda, Desta Yakob and Ke, Yu-Feng and Wang, Shou-Yan and Wen, Dong and Tong, Xiao-Guang and others},
  journal={Military Medical Research},
  volume={12},
  number={1},
  pages={14},
  year={2025},
  publisher={Springer}
}

@article{liu2026eeg,
  title={EEG-FM-Compass: Progress, Benchmarking, and Future Directions for EEG Foundation Models},
  author={Liu, Dingkun and Chen, Yuheng and Chen, Zhu and Cui, Zhenyao and Wen, Yaozhi and An, Jiayu and Luo, Jingwei and Wu, Dongrui},
  journal={National Science Review},
  pages={nwag466},
  year={2026},
  publisher={Oxford University Press}
}

@article{schirrmeister2017deep,
  title={Deep learning with convolutional neural networks for EEG decoding and visualization},
  author={Schirrmeister, Robin Tibor and Springenberg, Jost Tobias and Fiederer, Lukas Dominique Josef and Glasstetter, Martin and Eggensperger, Katharina and Tangermann, Michael and Hutter, Frank and Burgard, Wolfram and Ball, Tonio},
  journal={Human brain mapping},
  volume={38},
  number={11},
  pages={5391--5420},
  year={2017},
  publisher={Wiley Online Library}
}

@article{yao2025foundation,
  title={Foundation models for EEG decoding: Current progress and prospective research},
  author={Yao, Yuxuan and Wang, Hongbo and Chen, Li and Peng, Yiheng and Luo, Jingjing},
  journal={Journal of Neural Engineering},
  volume={22},
  number={6},
  pages={061002},
  year={2025},
  publisher={IOP Publishing}
}

@inproceedings{he2016deep,
  title={Deep residual learning for image recognition},
  author={He, Kaiming and Zhang, Xiangyu and Ren, Shaoqing and Sun, Jian},
  booktitle={Proceedings of the IEEE conference on computer vision and pattern recognition},
  pages={770--778},
  year={2016}
}

@article{rein2023gpqa,
  title={Gpqa: A graduate-level google-proof q\&a benchmark},
  author={Rein, David and Hou, Betty Li and Stickland, Asa Cooper and Petty, Jackson and Pang, Richard Yuanzhe and Dirani, Julien and Michael, Julian and Bowman, Samuel R},
  journal={arXiv preprint arXiv:2311.12022},
  year={2023}
}

@inproceedings{yue2024mmmu,
  title={Mmmu: A massive multi-discipline multimodal understanding and reasoning benchmark for expert agi},
  author={Yue, Xiang and Ni, Yuansheng and Zhang, Kai and Zheng, Tianyu and Liu, Ruoqi and Zhang, Ge and Stevens, Samuel and Jiang, Dongfu and Ren, Weiming and Sun, Yuxuan and others},
  booktitle={Proceedings of the IEEE/CVF conference on computer vision and pattern recognition},
  pages={9556--9567},
  year={2024}
}

@article{laurent2024lab,
  title={Lab-bench: Measuring capabilities of language models for biology research},
  author={Laurent, Jon M and Janizek, Joseph D and Ruzo, Michael and Hinks, Michaela M and Hammerling, Michael J and Narayanan, Siddharth and Ponnapati, Manvitha and White, Andrew D and Rodriques, Samuel G},
  journal={arXiv preprint arXiv:2407.10362},
  year={2024}
}

@article{singhal2023large,
  title={Large language models encode clinical knowledge},
  author={Singhal, Karan and Azizi, Shekoofeh and Tu, Tao and Mahdavi, S Sara and Wei, Jason and Chung, Hyung Won and Scales, Nathan and Tanwani, Ajay and Cole-Lewis, Heather and Pfohl, Stephen and others},
  journal={Nature},
  volume={620},
  number={7972},
  pages={172--180},
  year={2023},
  publisher={Nature Publishing Group UK London}
}

@inproceedings{hu2024omnimedvqa,
  title={Omnimedvqa: A new large-scale comprehensive evaluation benchmark for medical lvlm},
  author={Hu, Yutao and Li, Tianbin and Lu, Quanfeng and Shao, Wenqi and He, Junjun and Qiao, Yu and Luo, Ping},
  booktitle={2024 IEEE/CVF Conference on Computer Vision and Pattern Recognition (CVPR)},
  pages={22170--22183},
  year={2024},
  organization={IEEE}
}

@article{zuo2025medxpertqa,
  title={Medxpertqa: Benchmarking expert-level medical reasoning and understanding},
  author={Zuo, Yuxin and Qu, Shang and Li, Yifei and Chen, Zhangren and Zhu, Xuekai and Hua, Ermo and Zhang, Kaiyan and Ding, Ning and Zhou, Bowen},
  journal={arXiv preprint arXiv:2501.18362},
  year={2025}
}

@article{li2026hearts,
  title={Hearts: Benchmarking llm reasoning on health time series},
  author={Li, Sirui and Xiao, Shuhan and Joshi, Mihir and Metwally, Ahmed and McDuff, Daniel and Wang, Wei and Yang, Yuzhe},
  journal={arXiv preprint arXiv:2603.06638},
  year={2026}
}

@article{xiong2025eeg,
  title={EEG-FM-Bench: A Comprehensive Benchmark for the Systematic Evaluation and Diagnostic Analyses of EEG Foundation Models},
  author={Xiong, Wei and Li, Jiangtong and Li, Jie and Zhu, Kun and Jiang, Changjun},
  journal={arXiv preprint arXiv:2508.17742},
  year={2025}
}

@article{zhou2026brainbench,
  title={BrainBench: Benchmarking Large Language Models for Comprehensive EEG Understanding},
  author={Zhou, Yangxuan and Chen, Yuning and Wu, Chen and Wang, Jiquan and Li, Shijian and Pan, Gang and Zhao, Sha},
  journal={arXiv preprint arXiv:2608.04156},
  year={2026}
}

@article{team2026gemma,
  title={Gemma 4 technical report},
  author={Team, Gemma and Abd, Sherif El and Aggarwal, Vaibhav and Algayres, Robin and Andreev, Alek and Bachem, Olivier and Ballantyne, Ian and Brick, Cormac and C{\u{a}}rbune, Victor and Casbon, Michelle and others},
  journal={arXiv preprint arXiv:2607.02770},
  year={2026}
}

@article{yang2025qwen3,
  title={Qwen3 technical report},
  author={Yang, An and Li, Anfeng and Yang, Baosong and Zhang, Beichen and Hui, Binyuan and Zheng, Bo and Yu, Bowen and Gao, Chang and Huang, Chengen and Lv, Chenxu and others},
  journal={arXiv preprint arXiv:2505.09388},
  year={2025}
}

@misc{openai2025gpt41,
  author       = {{OpenAI}},
  title        = {Introducing GPT-4.1 in the API},
  year         = {2025},
  month        = apr,
  howpublished = {\url{https://openai.com/index/gpt-4-1/}},
  note         = {Accessed: 2026-10-04}
}

@article{xu2026deepseek,
  title={Deepseek-v4. 1-flash: Pushing the limits of kv cache compression},
  author={Xu, Anyi and Li, B and Lin, Bangcai and Xue, Bing and Xian, BingCheng and Xu, Bingzheng and Wu, Bochao and Zhang, Bowei and Deng, Boyi and Yu, CC and others},
  journal={arXiv preprint arXiv:2609.19969},
  year={2026}
}

@article{lai2026minimax,
  title={Minimax sparse attention},
  author={Lai, Xunhao and Xu, Weiqi and Yang, Yufeng and Chen, Qiaorui and Xu, Yang and Zeng, Lunbin and Li, Xiaolong and Sun, Haohai and Zhu, Haichao and Zhang, Vito and others},
  journal={arXiv preprint arXiv:2606.13392},
  year={2026}
}

@article{liu2026ministral,
  title={Ministral 3},
  author={Liu, Alexander H and Khandelwal, Kartik and Subramanian, Sandeep and Jouault, Victor and Rastogi, Abhinav and Sad{\'e}, Adrien and Jeffares, Alan and Jiang, Albert and Cahill, Alexandre and Gavaudan, Alexandre and others},
  journal={arXiv preprint arXiv:2601.08584},
  year={2026}
}

@misc{google2026gemini38flash,
  author       = {{Google}},
  title        = {Gemini 3.8 Flash},
  year         = {2026},
  howpublished = {\url{https://ai.google.dev/gemini-api/docs/models/gemini-3.8-flash}},
  note         = {Google AI for Developers. Accessed: 2026-10-04}
}

@misc{openai2026gpt56,
  author       = {{OpenAI}},
  title        = {How GPT-5.6 Fuses Frontier Intelligence with Frontier Efficiency},
  year         = {2026},
  month        = jul,
  howpublished = {\url{https://openai.com/index/gpt-5-6-frontier-intelligence-efficiency/}},
  note         = {Accessed: 2026-10-04}
}

@misc{qwen2026qwen35,
  title  = {{Qwen3.5}: Towards Native Multimodal Agents},
  author = {{Qwen Team}},
  month  = {February},
  year   = {2026},
  url    = {https://qwen.ai/blog?id=qwen3.5}
}

@misc{qwen2025qwen3,
  title         = {Qwen3 Technical Report},
  author        = {{Qwen Team}},
  year          = {2025},
  eprint        = {2505.09388},
  archivePrefix = {arXiv},
  primaryClass  = {cs.CL},
  url           = {https://arxiv.org/abs/2505.09388}
}

@misc{openai2022chatgpt,
  author       = {{OpenAI}},
  title        = {Introducing ChatGPT},
  year         = {2022},
  month        = nov,
  howpublished = {\url{https://openai.com/index/chatgpt/}},
  note         = {Accessed: 2026-10-04}
}

@article{bai2023qwen,
  title={Qwen technical report},
  author={Bai, Jinze and Bai, Shuai and Chu, Yunfei and Cui, Zeyu and Dang, Kai and Deng, Xiaodong and Fan, Yang and Ge, Wenbin and Han, Yu and Huang, Fei and others},
  journal={arXiv preprint arXiv:2309.16609},
  year={2023}
}

@misc{anthropic2024claude3,
  author       = {{Anthropic}},
  title        = {The Claude 3 Model Family: Opus, Sonnet, Haiku},
  year         = {2024},
  month        = mar,
  howpublished = {Claude 3 Model Card},
  note         = {Anthropic}
}

@misc{xai2023grok,
  author       = {{xAI}},
  title        = {Grok-1 Model Card},
  year         = {2023},
  month        = nov,
  howpublished = {\url{https://x.ai/news/grok/model-card}},
  note         = {Accessed: 2026-10-04}
}

@article{tangermann2012bci,
  title={Review of the BCI Competition IV},
  author={Tangermann, Michael and M{\"u}ller, Klaus-Robert and Aertsen, Ad
          and Birbaumer, Niels and Braun, Christoph and Brunner, Clemens
          and Leeb, Robert and others},
  journal={Frontiers in Neuroscience},
  volume={6},
  pages={55},
  year={2012},
  doi={10.3389/fnins.2012.00055}
}

@article{kappenman2021erpcore,
  title={ERP CORE: An Open Resource for Human Event-Related Potential Research},
  author={Kappenman, Emily S. and Farrens, Jaclyn L. and Zhang, Wendy
          and Stewart, Andrew X. and Luck, Steven J.},
  journal={NeuroImage},
  volume={225},
  pages={117465},
  year={2021},
  doi={10.1016/j.neuroimage.2020.117465}
}

@article{blankertz2006bci,
  title={The BCI Competition III: Validating Alternative Approaches to Actual BCI Problems},
  author={Blankertz, Benjamin and M{\"u}ller, Klaus-Robert and Krusienski, Dean J.
          and Schalk, Gerwin and Wolpaw, Jonathan R. and Schl{\"o}gl, Alois
          and Pfurtscheller, Gert and Mill{\'a}n, Jos{\'e} del R.
          and Schr{\"o}der, Michael and Birbaumer, Niels},
  journal={IEEE Transactions on Neural Systems and Rehabilitation Engineering},
  volume={14},
  number={2},
  pages={153--159},
  year={2006},
  doi={10.1109/TNSRE.2006.875642}
}

@article{zheng2015seed,
  title={Investigating Critical Frequency Bands and Channels for EEG-Based
         Emotion Recognition with Deep Neural Networks},
  author={Zheng, Wei-Long and Lu, Bao-Liang},
  journal={IEEE Transactions on Autonomous Mental Development},
  volume={7},
  number={3},
  pages={162--175},
  year={2015},
  doi={10.1109/TAMD.2015.2431497}
}

@article{zheng2019emotionmeter,
  title={EmotionMeter: A Multimodal Framework for Recognizing Human Emotions},
  author={Zheng, Wei-Long and Liu, Wei and Lu, Yifei and Lu, Bao-Liang
          and Cichocki, Andrzej},
  journal={IEEE Transactions on Cybernetics},
  volume={49},
  number={3},
  pages={1110--1122},
  year={2019},
  doi={10.1109/TCYB.2018.2797176}
}

@article{guillot2020dreem,
  title={Dreem Open Datasets: Multi-Scored Sleep Datasets to Compare
         Human and Automated Sleep Staging},
  author={Guillot, Antoine and Sauvet, Fabien and During, Emmanuel H.
          and Thorey, Valentin},
  journal={IEEE Transactions on Neural Systems and Rehabilitation Engineering},
  volume={28},
  number={9},
  pages={1955--1965},
  year={2020},
  doi={10.1109/TNSRE.2020.3011181}
}

@article{kemp2000sleep,
  title={Analysis of a Sleep-Dependent Neuronal Feedback Loop:
         The Slow-Wave Microcontinuity of the EEG},
  author={Kemp, Bob and Zwinderman, A. H. and Tuk, Bert
          and Kamphuisen, H. A. C. and Obery{\'e}, J. J. L.},
  journal={IEEE Transactions on Biomedical Engineering},
  volume={47},
  number={9},
  pages={1185--1194},
  year={2000},
  doi={10.1109/10.867928}
}

@article{guttag2010chbmit,
  author={Guttag, John},
  title={{CHB-MIT Scalp EEG Database}},
  journal={PhysioNet},
  year={2010},
  note={Version 1.0.0},
  doi={10.13026/C2K01R}
}

@article{accou2024sparrkulee,
  title={SparrKULee: A Speech-Evoked Auditory Response Repository from
         KU Leuven, Containing the EEG of 85 Participants},
  author={Accou, Bernd and Bollens, Lies and Gillis, Maaike
          and Verheijen, Wouter and Van Hamme, Hugo and Francart, Tom},
  journal={Data},
  volume={9},
  number={8},
  pages={94},
  year={2024},
  doi={10.3390/data9080094}
}

@article{gifford2022things,
  title={A Large and Rich EEG Dataset for Modeling Human Visual Object Recognition},
  author={Gifford, Alessandro T. and Dwivedi, Kshitij and Roig, Gemma
          and Cichy, Radoslaw M.},
  journal={NeuroImage},
  volume={264},
  pages={119754},
  year={2022},
  doi={10.1016/j.neuroimage.2022.119754}
}

@article{hollenstein2018zuco,
  title={ZuCo, a Simultaneous EEG and Eye-Tracking Resource for Natural Sentence Reading},
  author={Hollenstein, Nora and Rotsztejn, Jonathan and Troendle, Marius
          and Pedroni, Andreas and Zhang, Ce and Langer, Nicolas},
  journal={Scientific Data},
  volume={5},
  pages={180291},
  year={2018},
  doi={10.1038/sdata.2018.291}
}

@article{schalk2004bci2000,
  title={BCI2000: A General-Purpose Brain-Computer Interface (BCI) System},
  author={Schalk, Gerwin and McFarland, Dennis J. and Hinterberger, Thilo
          and Birbaumer, Niels and Wolpaw, Jonathan R.},
  journal={IEEE Transactions on Biomedical Engineering},
  volume={51},
  number={6},
  pages={1034--1043},
  year={2004},
  doi={10.1109/TBME.2004.827072}
}

@article{obeid2016tuh,
  title={The Temple University Hospital EEG Data Corpus},
  author={Obeid, Iyad and Picone, Joseph},
  journal={Frontiers in Neuroscience},
  volume={10},
  pages={196},
  year={2016},
  doi={10.3389/fnins.2016.00196}
}

@article{terzano2001cap,
  title={Atlas, Rules, and Recording Techniques for the Scoring of
         Cyclic Alternating Pattern (CAP) in Human Sleep},
  author={Terzano, Mario Giovanni and Parrino, Liborio and Sherieri, Adriano
          and Chervin, Ronald and Chokroverty, Sudhansu and Guilleminault, Christian
          and Hirshkowitz, Max and Mahowald, Mark and Moldofsky, Harvey
          and Rosa, Agostino and Thomas, Robert and Walters, Arthur},
  journal={Sleep Medicine},
  volume={2},
  number={6},
  pages={537--553},
  year={2001},
  doi={10.1016/S1389-9457(01)00149-6}
}

@inproceedings{liu2024eeg2video,
  title={{EEG2Video}: Towards Decoding Dynamic Visual Perception from EEG Signals},
  author={Liu, Xuan-Hao and Liu, Yan-Kai and Wang, Yansen and Ren, Kan
          and Shi, Hanwen and Wang, Zilong and Li, Dongsheng
          and Lu, Bao-Liang and Zheng, Wei-Long},
  booktitle={Advances in Neural Information Processing Systems},
  volume={37},
  pages={72245--72273},
  year={2024},
  doi={10.52202/079017-2306}
}

@article{lawhern2018eegnet,
  title={EEGNet: a compact convolutional neural network for EEG-based brain--computer interfaces},
  author={Lawhern, Vernon J and Solon, Amelia J and Waytowich, Nicholas R and Gordon, Stephen M and Hung, Chou P and Lance, Brent J},
  journal={Journal of neural engineering},
  volume={15},
  number={5},
  pages={056013},
  year={2018},
  publisher={iOP Publishing}
}

@article{song2022eeg,
  title={EEG conformer: Convolutional transformer for EEG decoding and visualization},
  author={Song, Yonghao and Zheng, Qingqing and Liu, Bingchuan and Gao, Xiaorong},
  journal={IEEE Transactions on Neural Systems and Rehabilitation Engineering},
  volume={31},
  pages={710--719},
  year={2022},
  publisher={IEEE}
}

@article{wilson2025deep,
  title={Deep riemannian networks for end-to-end eeg decoding},
  author={Wilson, Daniel and Schirrmeister, Robin T and Gemein, Lukas AW and Ball, Tonio},
  journal={Imaging Neuroscience},
  volume={3},
  pages={imag\_a\_00511},
  year={2025},
  publisher={MIT Press 255 Main Street, 9th Floor, Cambridge, Massachusetts 02142, USA~…}
}

@article{tibermacine2026stiefel,
  title={Stiefel-SPD manifold graph convolution for end-to-end EEG learning},
  author={Tibermacine, Imad Eddine and Russo, Samuele and Napoli, Christian},
  journal={IEEE Transactions on Neural Systems and Rehabilitation Engineering},
  volume={34},
  pages={595--606},
  year={2026},
  publisher={IEEE}
}

@inproceedings{cui2024neuro,
  title={Neuro-gpt: Towards a foundation model for eeg},
  author={Cui, Wenhui and Jeong, Woojae and Th{\"o}lke, Philipp and Medani, Takfarinas and Jerbi, Karim and Joshi, Anand A and Leahy, Richard M},
  booktitle={2024 IEEE International Symposium on Biomedical Imaging (ISBI)},
  pages={1--5},
  year={2024},
  organization={IEEE}
}

@inproceedings{mishra2025thought2text,
  title={Thought2Text: Text generation from EEG signal using large language models (LLMs)},
  author={Mishra, Abhijit and Shukla, Shreya and Torres, Jose and Gwizdka, Jacek and Roychowdhury, Shounak},
  booktitle={Findings of the association for computational linguistics: NAACL 2025},
  pages={3747--3759},
  year={2025}
}

@article{chen2026mindalign,
  title={MindAlign: Bridging EEG, Vision, and Language for Zero-Shot Visual Decoding},
  author={Chen, Zexuan and Liu, Sichao and Lu, Runhao and Qi, Huichao and Woolgar, Alexandra and Wang, Xi Vincent and Wang, Lihui},
  journal={arXiv preprint arXiv:2605.24523},
  year={2026}
}

@article{qiu2026eeg,
  title={EEG-VLM: A Hierarchical Vision-Language Model With Multi-Level Feature Alignment and Visually Enhanced Language-Guided Reasoning for EEG Image-Based Sleep Stage Prediction},
  author={Qiu, Xihe and Ma, Gengchen and Wang, Haoyu and Zhan, Chen and Tan, Xiaoyu and Li, Shuo},
  journal={IEEE Journal of Biomedical and Health Informatics},
  year={2026},
  publisher={IEEE}
}

\appendix

\section{Examples of Open-Ended Reference Answers and Scoring Rubrics}
\label{app:rubric_examples}

This section provides two complete examples of the reference answers and
item-specific scoring rubrics used for open-ended questions in EEGProbe.
The examples are drawn from Frequency-Band Analysis (FBA) and Brain-Region
Interpretation (BRI), respectively.

\subsection{Frequency-Band Analysis Example}

\noindent\textbf{Example ID:} \texttt{bandalpha\_001}

\noindent\textbf{Reference answer.}
Signal A has stronger posterior 8--13 Hz alpha-band activity. The posterior
band-power difference in the model-visible window is approximately 11.9 dB,
with P3, P4, and P8 supporting the same direction and the change concentrated
over posterior regions. This pattern is consistent with stronger posterior
alpha during eye closure and alpha attenuation during eye opening
(alpha blocking/Berger effect). Because acquisition order was fixed, this is
treated as a condition-level association and does not exclude effects of order,
vigilance, or other recording factors.

\begin{table}[h]
	\centering
	\caption{Scoring rubric for the FBA example \texttt{bandalpha\_001}.}
	\small
	\begin{tabularx}{\linewidth}{X c}
		\toprule
		\textbf{Criterion} & \textbf{Weight} \\
		\midrule
		Correctly identifies Signal A as having stronger posterior alpha activity.
		& 0.30 \\
		
		Reports posterior regions or supporting channels consistent with the
		evidence, such as P3, P4, and P8.
		& 0.25 \\
		
		Characterizes the effect as strong and correctly describes its posterior
		specificity.
		& 0.25 \\
		
		Relates the pattern conservatively to alpha blocking or the Berger effect.
		& 0.20 \\
		\midrule
		\textbf{Total} & \textbf{1.00} \\
		\bottomrule
	\end{tabularx}
\end{table}

The reference answer is supported by a posterior band-power difference of
11.8951 dB, consistent directional effects at P3, P4, and P8, agreement
between the 19-channel and 64-channel analyses, and a whole-record stability
value of 1.0.

\subsection{Brain-Region Interpretation Example}

\noindent\textbf{Example ID:} \texttt{rsni\_v2\_035}

\noindent\textbf{Reference answer.}
The sample contains a late positive activity pattern with relative
centroparietal prominence. From 300--600 ms, the centroparietal mean is
37.4~$\mu$V and exceeds the strongest comparison region, the central region
at 28.8~$\mu$V, by approximately 8.7~$\mu$V. The regional trace peaks near
371 ms at 60.2~$\mu$V, with P3, Pz, and CPz providing the clearest channel
support. This stimulus-locked spatiotemporal pattern is compatible with a
P3-like process related to target evaluation or context updating. A single
scalp epoch, however, does not establish a definitive component identity,
precise cortical source, or deterministic cognitive state.

\begin{table}[h]
	\centering
	\caption{Scoring rubric for the BRI example \texttt{rsni\_v2\_035}.}
	\small
	\begin{tabularx}{\linewidth}{X c}
		\toprule
		\textbf{Criterion} & \textbf{Weight} \\
		\midrule
		Correctly identifies a late positive pattern with relative centroparietal
		prominence.
		& 0.25 \\
		
		Reports positive polarity and a compatible peak latency near 371 ms
		($\pm$50 ms).
		& 0.20 \\
		
		Uses sample-specific evidence such as P3, Pz, and CPz and the
		approximately 8.7~$\mu$V centroparietal dominance.
		& 0.25 \\
		
		Provides a conservative P3-like target-evaluation or context-updating
		interpretation.
		& 0.20 \\
		
		Avoids definitive claims about component identity, source localization,
		or cognitive state.
		& 0.10 \\
		\midrule
		\textbf{Total} & \textbf{1.00} \\
		\bottomrule
	\end{tabularx}
\end{table}

The reference answer is supported by a regional difference of
8.6628~$\mu$V, a peak latency of 371.09 ms, a regional peak of
60.195~$\mu$V, and a positive-channel proportion of 1.0.

\end{document}